\documentclass{magicstyle/magic}

\usepackage{amsmath,amssymb,amsfonts}
\usepackage{xspace}
\usepackage{enumitem}
\usepackage{makecell}
\usepackage{pifont}
\usepackage{array}
\usepackage{colortbl}
\setlist[itemize]{leftmargin=*,itemsep=0em,parsep=0.3em,topsep=0.3em}

\usepackage{tikz}
\usetikzlibrary{positioning,fit,arrows.meta,backgrounds,patterns,calc,decorations.pathreplacing}

\usepackage{amsthm}
\theoremstyle{plain}

\theoremstyle{definition}

\theoremstyle{remark}

\newcommand{\moe}{MoFE}
\newcommand{\Lfm}{\mathcal{L}_{\mathrm{FM}}}

\let\cite\citep

\DeclareTextFontCommand{\tabtextbf}{\bfseries}
\AtBeginEnvironment{tabular}{\let\textbf\tabtextbf}

\crefname{figure}{Figure}{Figures}
\crefname{table}{Table}{Tables}
\crefname{section}{Section}{Sections}
\crefname{subsection}{Section}{Sections}
\crefname{subsubsection}{Section}{Sections}
\crefname{appendix}{Appendix}{Appendices}
\crefname{equation}{Equation}{Equations}

\title{%
  {\fontsize{18pt}{22pt}\selectfont
   \textcolor{magicblue}{Force2Act: A Force-Aware Visuomotor Policy\\for Generalizable Compliant Robot Manipulation}}%
}
\author[1]{Keerthi Vasan Murugan}
\author[1,2]{Fangguo Zhao}
\author[1]{Jiafei Duan*}

\affiliation[1]{National University of Singapore}
\affiliation[2]{Zhejiang University}

\contribution[*]{Corresponding Author}

\abstract{Contact-rich manipulation requires a robot to regulate the force it applies, not only where
it moves.
Existing force-aware policies learn this force from demonstrations and therefore reproduce
the force levels they were trained on, which fails once the interaction demands more force
than any demonstration.
We present Force2Act, a hybrid force--position policy built on two complementary mechanisms.
A global force policy plans the motion and the interaction force once per action chunk, with
force represented relative to the measured interaction, so that the planned force scales with
the load.
Between planning steps, a reactive compliance policy, termed Mixture of Force Experts,
identifies at every control step the force expert that governs the interaction and regulates
the force through it, so that the contact holds when it changes within a chunk.
We further introduce Force2Act-Bench, a simulation benchmark of five contact-rich tasks that
gives the community a common and repeatable way to test force-aware policies and measure how
well they regulate the interaction force.
Force2Act achieves an overall success rate of 93.8\%, compared with 77.0\% for the
strongest baseline, and its reactive regulation reduces the failures on block pivoting from
30 to 10 of 150 trials.
On a real robot, Force2Act reaches an average success rate of 94.2\% over four tasks,
against at most 27.5\% for the baselines.
}

\newcommand{\github}{\raisebox{-1.5pt}{\includegraphics[height=1.05em]{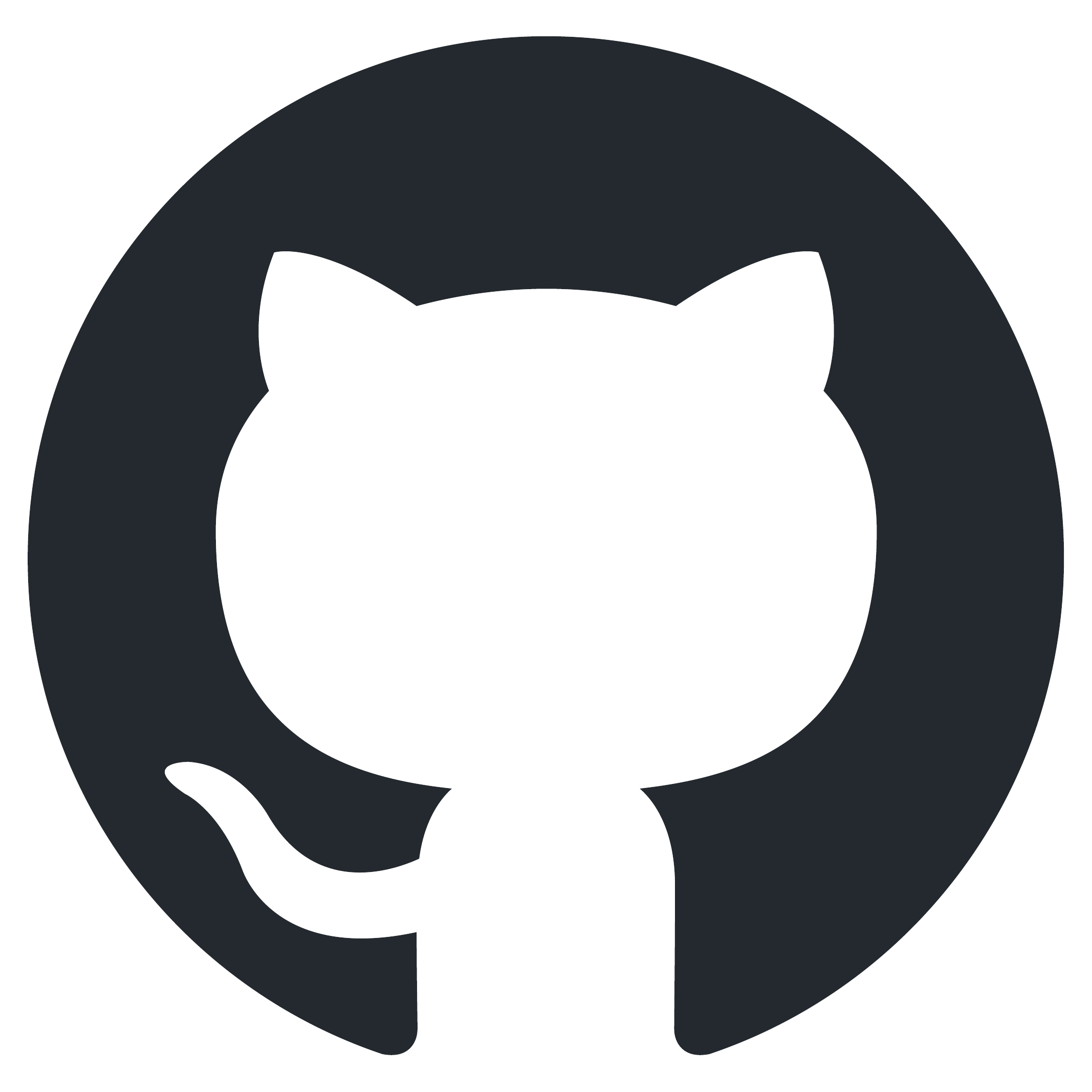}}\xspace}
\newcommand{\huggingface}{\raisebox{-1.5pt}{\includegraphics[height=1.05em]{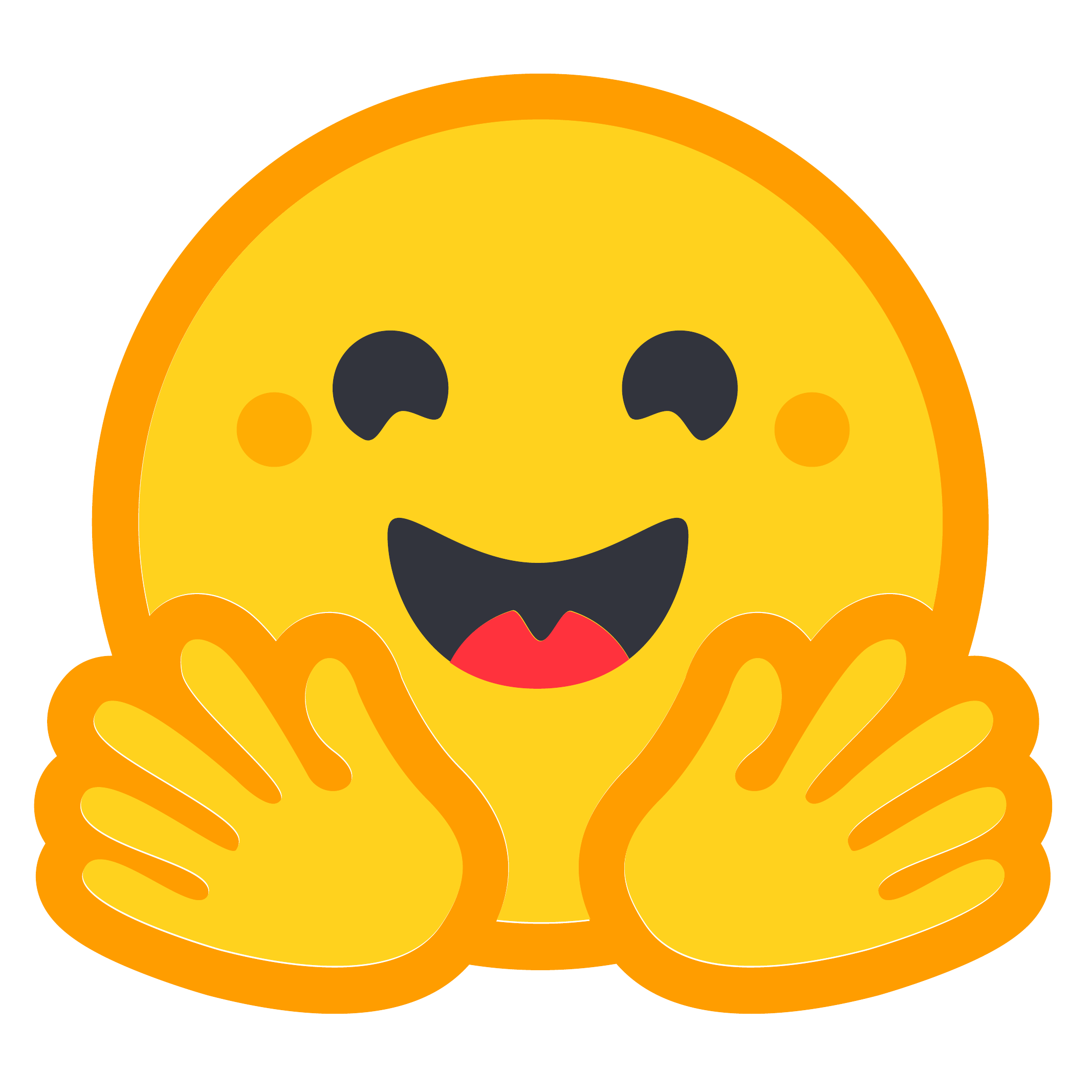}}\xspace}
\newcommand{\hfdataset}{\raisebox{-1.5pt}{\includegraphics[height=1.05em]{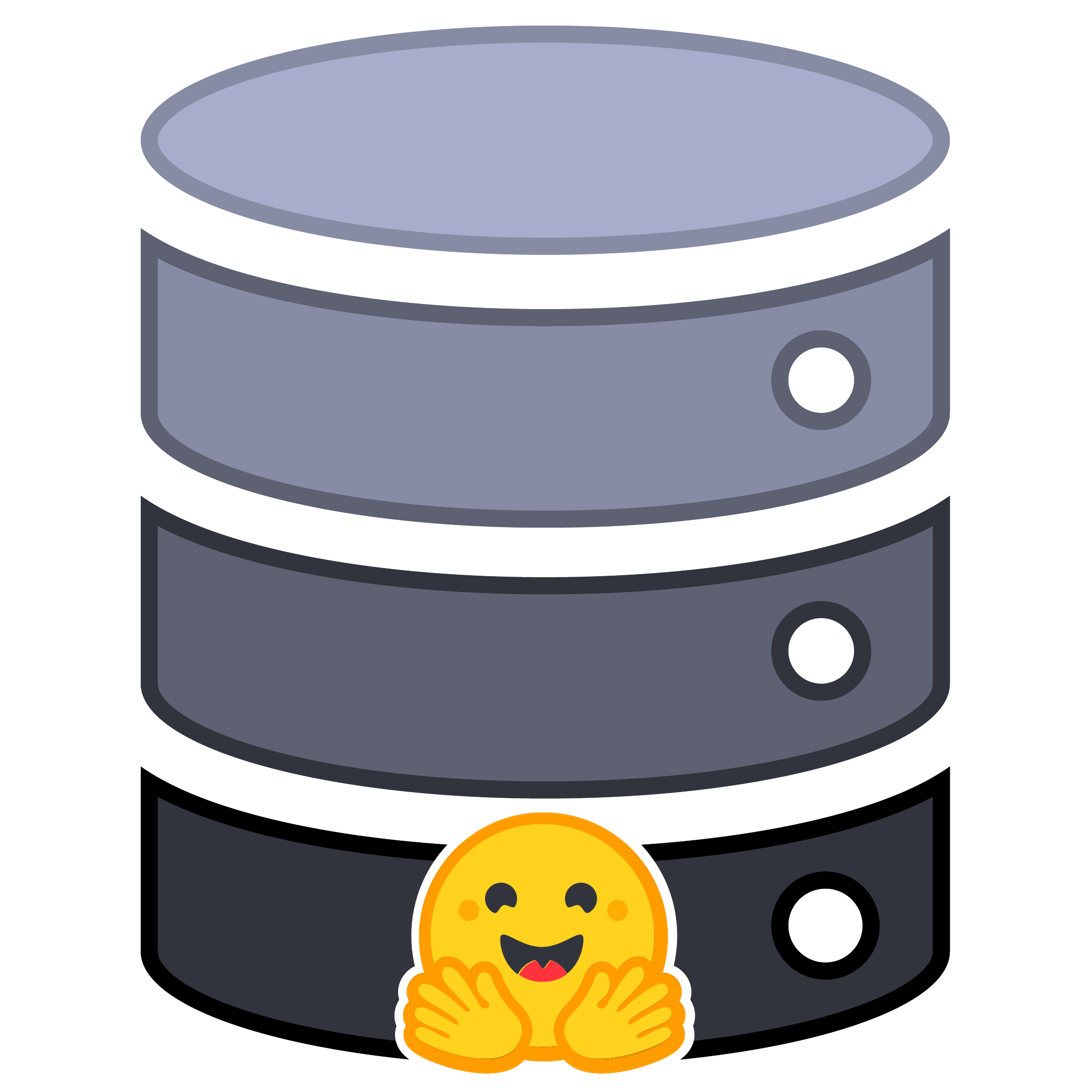}}\xspace}
\newcommand{\emailLogo}{\raisebox{-1.5pt}{\includegraphics[height=1.05em]{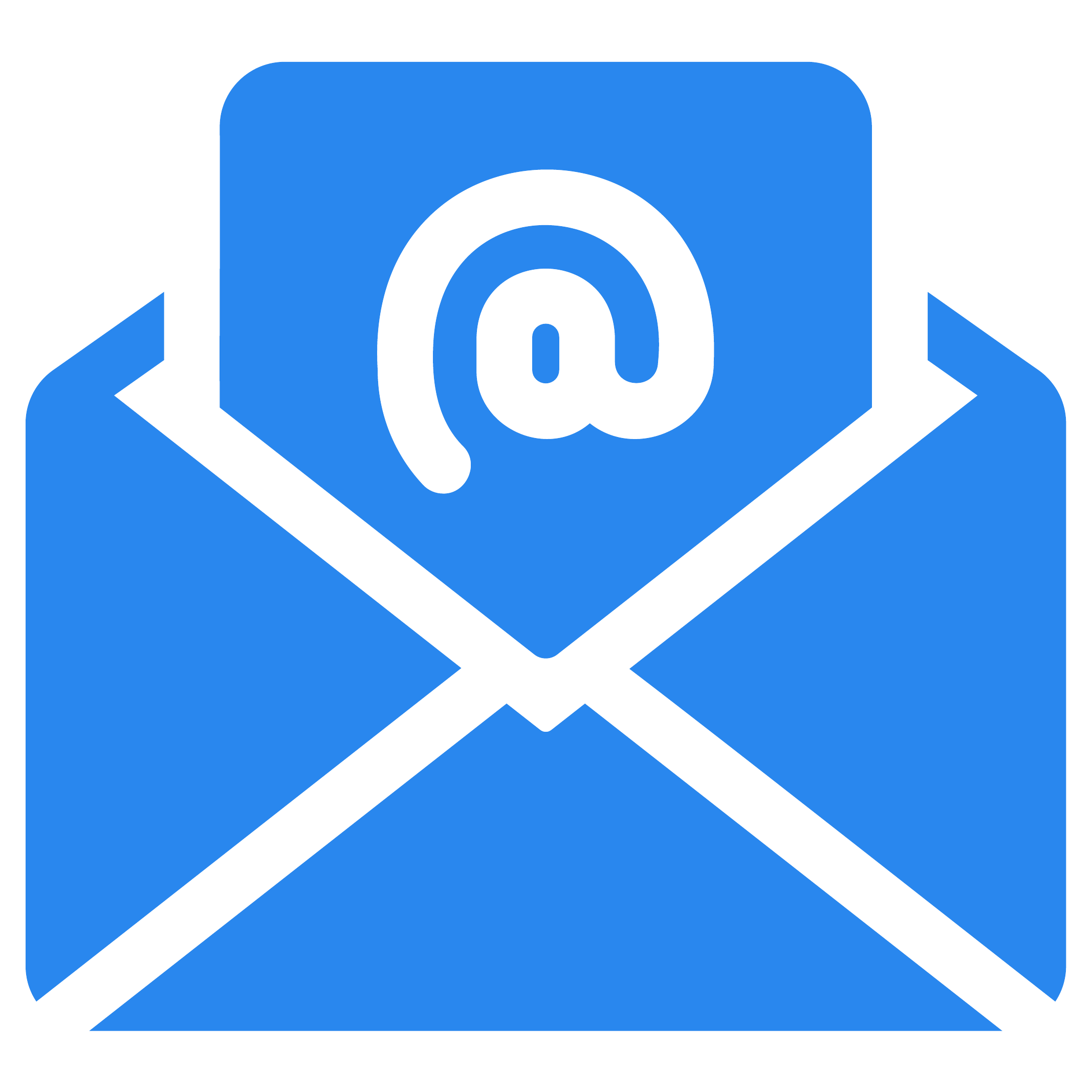}}\xspace}
\newcommand{\websiteLogo}{\raisebox{-1.5pt}{\includegraphics[height=1.05em]{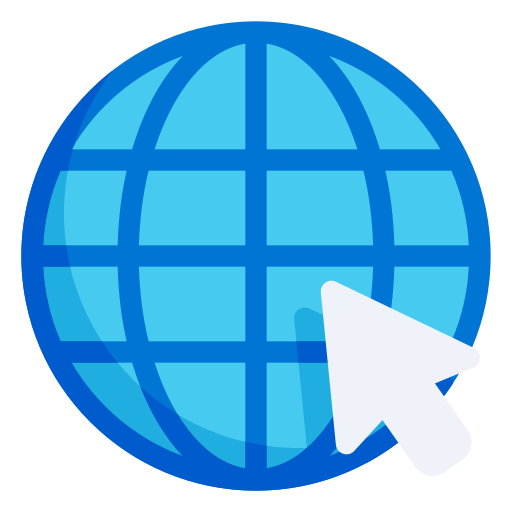}}\xspace}
\metadata[\websiteLogo Webpage:]{\href{https://magiclab-nus.github.io/Force2Act/}{\texttt{magiclab-nus.github.io/Force2Act}}}
\metadata[\vspace{.3em}\github Code:]{\href{https://github.com/MAGICLAB-NUS/Force2Act}{\texttt{github.com/MAGICLAB-NUS/Force2Act}}}
\metadata[\vspace{.3em}\huggingface Models:]{\href{https://huggingface.co/NUSMAGIC/Force2Act-Sim}{\texttt{huggingface.co/NUSMAGIC/Force2Act-Sim}}}
\metadata[\vspace{.3em}\hfdataset Dataset:]{\href{https://huggingface.co/datasets/NUSMAGIC/Force2Act-Bench-Data}{\texttt{huggingface.co/datasets/NUSMAGIC/Force2Act-Bench-Data}}}
\metadata[\vspace{.3em}\emailLogo Contact:]{\email{duanj1@nus.edu.sg}}

\begin{document}

\maketitle

% Teaser figure (content from the CVPR version), shown below the title block.
\begin{figure}[!ht]
  \centering
  \includegraphics[width=\textwidth]{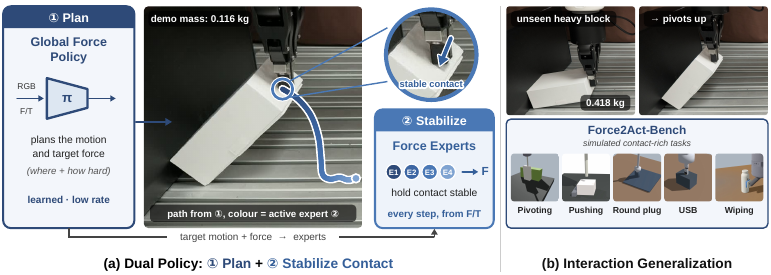}
  \caption{\textbf{Regulating force for contact-rich manipulation.}
  (a) The global force policy plans the motion and the target force once per action chunk,
  and force experts keep the contact stable at every step from the measured force.
  The path of a real rollout is coloured by the active expert.
  (b) Force2Act pivots a 0.418\,kg block after demonstrations with a 0.116\,kg block, and is
  evaluated on the five simulated tasks of Force2Act-Bench.}
  \label{fig:teaser}
\end{figure}

%%%%%%%%% MAIN PAPER %%%%%%%%%
\section{Introduction}
\label{sec:intro}

Humans manipulate the physical world through a continuous interplay of perception, action, and physical feedback~\cite{westling1984factors}. When wiping a surface, inserting a connector, or assembling parts, we do more than guide our hands to the right location: we adjust how firmly we press, how readily we yield to resistance, and how we move as contact changes. These adjustments are essential because the same motion can produce different outcomes depending on the object's weight, surface friction, and contact geometry. Enabling robots to perform such contact-rich manipulation therefore requires coupling \emph{where} they move with \emph{how} they apply and adapt force throughout an interaction.

Visuomotor policies~\cite{chi2023diffusion} and vision-language-action (VLA) models~\cite{black2024pi0,intelligence2025pi05,DBLP:conf/icra/LeeDFDLLFZWLHPWHFVFFK26,DBLP:journals/corr/abs-2605-02881} have greatly advanced general manipulation by mapping visual observations and task instructions to robot actions. However, policies that rely primarily on vision and predict motion alone leave contact forces largely implicit. Visually similar interactions can involve different loads and resistance, making images alone insufficient to determine whether the robot should press harder, yield, or change its motion. Force-aware policies address this gap by incorporating force observations~\cite{he2025foar,yu2025forcevla}, predicting force as an auxiliary target~\cite{zhang2025tavla,zhang2026forceflow}, introducing fast feedback loops~\cite{xue2025rdp,li2026favla,vanjani2026damvla}, or explicitly commanding force and compliance~\cite{fang2026forcepolicy,li2026forcevla2,hou2025adaptive,kim2026compvla}.

Yet sensing force does not by itself establish how to adapt the interaction when physical conditions change. Approaches that use force as an additional observation or auxiliary training target~\cite{yu2025forcevla,chen2026implicitrdp} provide information about contact without explicitly specifying how the commanded force should respond. Approaches that predict force or stiffness from demonstrations~\cite{li2026forcevla2,hou2025adaptive} offer more direct control, but supervision from recorded values alone does not ensure that the learned response remains appropriate under unfamiliar loads or contact conditions. The central challenge is therefore to use physical feedback to determine the force and compliance required by the \emph{current} interaction, including when its dynamics differ from those encountered during training. We refer to this capability as \emph{interaction generalization}. Consider a robot pressing on a box from above while pushing it along an incline whose slope gradually increases (\cref{fig:method}). As the incline steepens, the robot must adapt its applied force to maintain controlled motion, even though images may reveal little about the changing load before the box begins to slip. Insufficient force can allow the box to slide back, while excessive pressing can increase resistance and stall its motion. Success therefore requires more than recognizing the box and following a trajectory: the robot must relate the measured interaction to the force needed to sustain the task.

To address this, we propose \textbf{Force2Act} (\cref{fig:teaser}), a hybrid force--position policy designed for interaction generalization by coupling global motion and force planning with reactive force regulation. Once per action chunk, its global force policy generates future poses and wrenches relative to the measured interaction using a flow-matching diffusion transformer~\cite{lipman2023flow,peebles2023dit}. Since contact conditions can change before the next planning step, for instance when a block starts to tip, we introduce \textbf{Mixture of Force Experts (\moe{})}, a reactive compliance policy that regulates force at every control step to maintain the contact conditions on which the planned force depends. It selects the force expert corresponding to the current interaction and evaluates it using the latest measurement, such as a sticking expert that increases normal force as the tangential load grows~\cite{DBLP:books/daglib/0022548}. To evaluate this ability to generalize across physical interactions, we introduce \textbf{Force2Act-Bench}, a simulation benchmark~\cite{genesis2024} comprising five contact-rich tasks designed to assess how well a policy understands and regulates interaction forces beyond reproducing its demonstrations. On Force2Act-Bench, Force2Act achieves an overall success rate of 93.8\%, compared with 77.0\% for the strongest baseline. On a real robot, Force2Act reaches an average success rate of 94.2\% over four tasks, against at most 27.5\% for the baselines.

\section{Related Work}
\label{sec:related}

% Two families: visuomotor policies (DP, ManiFlow, RT-1, Octo; Force2Act belongs here)
% vs. VLAs built on a pretrained VLM (RT-2, OpenVLA, pi0, ...). Shared limitation: no force.
\paragraph{Visuomotor policies and vision-language-action models.}
Visuomotor policies~\cite{chi2023diffusion,yan2025maniflow,brohan2023rt1,octo2024} are
trained on robot demonstrations with task-specific or lightly pretrained encoders.
Vision-language-action (VLA)
models~\cite{brohan2023rt2,kim2024openvla,kim2025openvlaoft,black2024pi0,intelligence2025pi05,bjorck2025gr00t,DBLP:conf/icra/LeeDFDLLFZWLHPWHFVFFK26,DBLP:journals/corr/abs-2605-02881}
instead build on a pretrained vision-language model, transferring web-scale semantic priors
to robotic control.
Both visuomotor policies and VLAs condition the policy on vision, language, and
proprioception and output kinematic targets, a formulation in which interaction forces are
neither observed nor regulated.

% Force-aware policies (main): force as observation -> timescales -> explicit regulation -> our difference.
\paragraph{Force-aware manipulation policies.}
Force-aware visuomotor policies and VLAs differ mainly in the role that force plays.
A first group uses force as an additional observation.
FoAR~\cite{he2025foar} modulates force features according to a predicted contact
probability.
Among VLAs, ForceVLA~\cite{yu2025forcevla} integrates force into the action expert of
$\pi_0$ through a mixture-of-experts module.
TA-VLA~\cite{zhang2025tavla} additionally predicts future torques jointly with its actions,
and ForceFlow~\cite{zhang2026forceflow} predicts the contact force of the next step, in both
cases only as an auxiliary training objective that is not used during execution.
A second group addresses the mismatch between the rate of visual planning and that
of contact dynamics.
RDP~\cite{xue2025rdp} augments a slow latent diffusion policy with a fast
force-conditioned module, whereas ImplicitRDP~\cite{chen2026implicitrdp} realizes both
timescales within a single network.
On the VLA side, FAVLA~\cite{li2026favla} adapts its execution frequency to the predicted
force variation, while DAM-VLA~\cite{vanjani2026damvla} processes each modality
asynchronously at its native rate.
In both groups, force shapes what the policy perceives and anticipates, but the
interaction force itself is not regulated during execution.
A third group regulates the interaction explicitly.
ACP~\cite{hou2025adaptive} predicts the stiffness of an impedance
controller~\cite{hogan1985impedance} from force feedback, and
CompVLA~\cite{kim2026compvla} predicts it within a VLA from vision and language.
TacDiffusion~\cite{wu2025tacdiffusion} formulates the policy in the force domain, and
ForceMimic~\cite{liu2025forcemimic} learns force-centric skills from force-motion capture.
Force Policy~\cite{fang2026forcepolicy} employs hybrid force--position
control~\cite{raibert1981hybrid} in a learned interaction frame, while
ForceVLA2~\cite{li2026forcevla2} conditions hybrid control on force-aware task prompts.
These methods apply appropriate forces under conditions that resemble their
demonstrations, but the commanded force or stiffness is learned from the recorded values,
so it does not adapt when the interaction dynamics leave the training distribution.

% Physical structure and mixtures of experts -> our experts are physical force laws, routed by geometry.
\paragraph{Physical structure and mixtures of experts.}
Embedding physical structure in a learned model can improve its generalization beyond the
training data.
Hamiltonian neural networks~\cite{greydanus2019hamiltonian}, sparse identification of governing
equations~\cite{brunton2016sindy}, and residual
physics~\cite{zeng2019tossingbot,johannink2019residual} each embed a single physical model,
whereas a contact-rich task passes through several contacts that are governed by different laws.
Mixtures of experts~\cite{jacobs1991adaptive,shazeer2017outrageously} choose among several
specialized models through a learned router.
Robot policies employ them within VLAs~\cite{zhou2025chatvla,zhao2025more} and, in
ForceVLA~\cite{yu2025forcevla} and ForceVLA2~\cite{li2026forcevla2}, to fuse force with the
other modalities.
Force2Act combines both ideas, with physical force laws as experts whose admissibility is
decided by the contact geometry, and a router learned from the demonstrated force alone
through the Gumbel-softmax relaxation~\cite{DBLP:conf/iclr/JangGP17,DBLP:conf/iclr/MaddisonMT17}.

\section{Method}
\label{sec:method}

\begin{figure*}[!tbp]
  \centering
  \includegraphics[width=\textwidth]{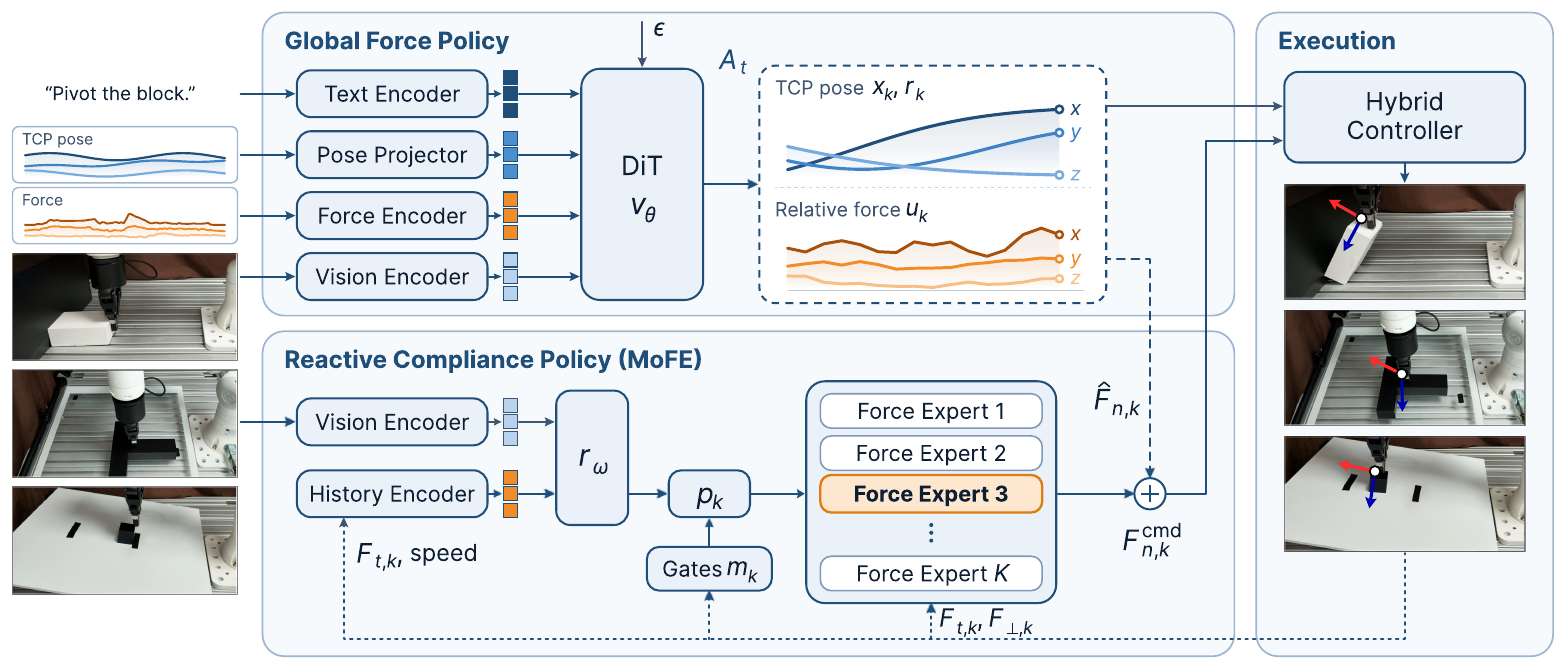}
  \caption{\textbf{Overview of Force2Act.}
    The global force policy plans poses and wrenches once per action chunk from images,
    language, and force and pose histories.
    At every control step, the reactive compliance policy, \moe{}, regulates the force.
    Geometric gates exclude the force experts that cannot apply, a router selects one of the
    remaining experts, and the selected expert sets the normal force along the $z$-axis of the
    motion frame.}
  \label{fig:method}
\end{figure*}

Force2Act separates contact-rich manipulation into planning the interaction and
regulating it (\cref{fig:method}).
A \emph{global force policy} plans the motion and the evolution of the interaction force
once per action chunk, and a \emph{reactive compliance policy}, \moe{}, sets the normal
force at every control step.
Both policies are executed by a hybrid force--position controller~\cite{raibert1981hybrid},
which regulates the translational force along the contact normal and tracks the planned
pose in the remaining directions.

% ---------------------------------------------------------------------------
\subsection{Global Force Policy}
\label{sec:method:policy}

At control step $t$, the global force policy receives short histories of camera images
$\mathcal{I}_t$, end-effector poses $\mathcal{P}_t$, and wrench measurements
$\mathcal{W}_t = (w_{t-n+1}, \dots, w_t)$ with $w = (f, m) \in \mathbb{R}^6$ from a wrist
force/torque sensor, together with a language instruction $\ell$.
It predicts an action chunk~\cite{DBLP:conf/rss/ZhaoKLF23} $A_t = (a_t, \dots, a_{t+H-1})$, of which
the first $H_a < H$ actions are executed before the policy is queried again.
Each action $a_k = (x_k, r_k, u_k)$ comprises an end-effector position
$x_k \in \mathbb{R}^3$, an orientation $r_k \in \mathbb{R}^6$ in the continuous 6D
representation~\cite{zhou2019continuity}, and a wrench $u_k \in \mathbb{R}^6$.

Force enters the policy in relative units.
Absolute force magnitudes vary with the object and the contact, and a policy that
regresses them reproduces the magnitudes of its demonstrations.
We therefore normalize the observed and predicted wrenches by the recent force level
\begin{equation}
  s_t = \max\Big(\tfrac{1}{n}\textstyle\sum_{k=t-n+1}^{t} \|f_k\|,\; s_{\min}\Big),
  \label{eq:scale}
\end{equation}
where the lower bound $s_{\min}$ prevents sensor noise in free space from being
amplified.
The policy observes $\mathcal{W}_t / s_t$ and predicts relative wrenches $u_k$, which are
converted to physical units as $\hat{w}_k = s_t\, u_k$.
Since input and output share one scale, the policy determines the direction and timing
of the force, while its magnitude is rescaled by the measurement through $s_t$, so the planned
force grows with the measured load although the policy is trained at a single load.

Modality-specific encoders $E_\phi$ (Appendix~\ref{sec:implementation}) turn the observation
into context tokens that condition the action head $v_\theta$, a diffusion
transformer~\cite{peebles2023dit} trained by conditional flow matching~\cite{lipman2023flow}
and integrated with a few Euler steps at inference.

% ---------------------------------------------------------------------------

\subsection{Reactive Compliance Policy}
\label{sec:method:reactive}

The global force policy is queried only once every $H_a$ steps, so its planned force is
executed open loop for $H_a = 8$ steps (0.53\,s).
When the contact changes within a chunk, for instance as a block starts to tip, the planned
force can under-press, and the contact is lost.
The reactive compliance policy therefore regulates the force between planning steps through a
force expert that is evaluated on the current measurement at every control step.
It is realized as Mixture of Force Experts (\moe{}), whose experts are physical force laws with
constants learned from demonstrations (\cref{sec:method:experts}), selected by geometric
gates and a learned router (\cref{sec:method:routing}), and trained jointly from the
demonstrated normal force (\cref{sec:method:training}).

\subsubsection{Force Experts}
\label{sec:method:experts}

\begin{figure}[!tbp]
  \centering
  \input{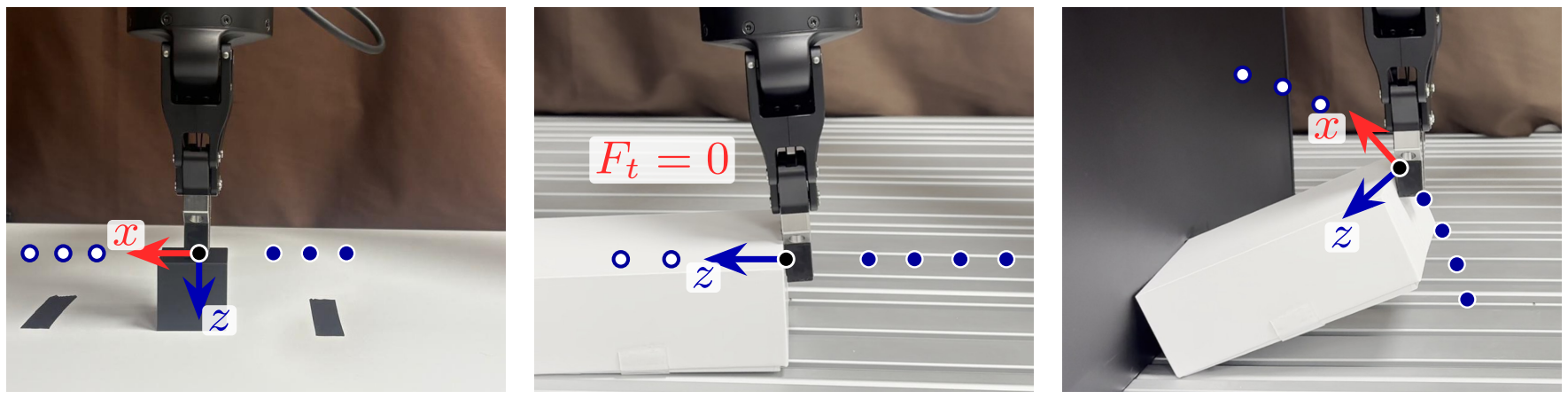}
  \caption{Motion frame in sliding, pushing, and pivoting. The $x$-axis follows the
  end-effector trajectory (filled for past and hollow for planned points) and the $z$-axis is the
  force-controlled direction.}
  \label{fig:motion_frame}
\end{figure}

We decompose the contact force into a normal component $F_n$, which the robot commands,
and a tangential component $F_t$, which results from the interaction.
Both are expressed in a motion frame (\cref{fig:motion_frame}), which is constructed from
how the end-effector moves.
Its $x$-axis follows the measured motion and its $z$-axis the component of the measured
force orthogonal to it, so that $F_t$ acts along the motion and $F_n$ across it.
When the end-effector is nearly at rest, or when the force is nearly parallel to the motion,
as in pushing, $z$ is taken along the force itself, with $F_t = 0$.
% We further denote by $F_\perp$ the magnitude of the force component orthogonal to the
% tool axis.
For an insertion, the observed load is instead $F_\perp$, the sideways force orthogonal to the
insertion axis.

The experts of \moe{} follow from Coulomb friction~\cite{DBLP:books/daglib/0022548}, under which
a contact transmits a tangential force only up to $F_t \le \mu F_n$.
This condition distinguishes four contact cases in the motion frame, which
\cref{tab:experts} lists with the load $\psi_j$ that each case must resist, the measured
quantity on which the normal force of its expert $j$ depends, and the condition under which
it applies.
In a \emph{sticking} contact, the end-effector presses on an object and moves it without
slipping, so the normal force must grow with the tangential load, $F_n \ge F_t / \mu$.
Its load is therefore the tangential force ($\psi_j = F_t$), and the case applies whenever $F_t > 0$.
In a \emph{sliding} contact, the end-effector slips along the surface it touches, as in
wiping, and the tangential force $F_t = \mu F_n$ is caused by the normal force rather than
resisted by it, so the normal force is prescribed and does not depend on a load
($\psi_j = 0$).
In a \emph{pushing} contact, the force acts along the motion, so $F_t = 0$ by the
construction of the motion frame, and the contact carries a steady force ($\psi_j = 0$).
In an \emph{insertion}, the force acts along the tool axis, and the axial push must also
overcome the friction that arises when the plug presses sideways against the wall, so its
load is the lateral force ($\psi_j = F_\perp$).
In every case, the normal force is affine in the load, and each expert takes the form
\begin{equation}
  f_j = c_j\, \psi_j + b_j, \qquad
  \psi_j \in \{0,\; F_t,\; F_\perp\},
  \label{eq:experts}
\end{equation}
where $\psi_j$ is the load listed in \cref{tab:experts}.
The law of a case fixes the form of its experts but not their constants, which can differ
between contacts of the same case, so each case has one or more experts.
We use two sticking experts and three sliding experts, which gives $K = 7$ experts, and
the router learns during training which contacts each expert covers
(\cref{sec:method:training}).
% anchored late and kept out of this column, so that Figure 2 has text below it and these floats open the next page
\suppressfloats[t]%
\begin{table}[!tbp]
  \centering
  \footnotesize
  \caption{\textbf{Contact cases of \moe{}.}
    Each expert predicts $F_n = c_j \psi_j + b_j$ from the load $\psi_j$ of its contact
    case and is admissible only while the condition of the case holds.
    The experts of a case share its form but learn separate constants.}
  \label{tab:experts}
  \renewcommand{\arraystretch}{1.1}
  \setlength{\tabcolsep}{4pt}
  \begin{tabular}{l c c >{\raggedright\arraybackslash}p{3.6cm}}
    \toprule
    Contact case & Experts & Load $\psi_j$ & Admissible when \\
    \midrule
    Sticking & 2 & $F_t$ & $F_t > 0$ \\
    Sliding & 3 & $0$ & always \\
    Pushing & 1 & $0$ & the force is along the motion \\
    Insertion & 1 & $F_\perp$ & the force is along the tool axis \\
    \bottomrule
  \end{tabular}
\end{table}
The slope $c_j$ and the offset $b_j$ are the only parameters of an expert, and no friction
coefficient, mass, or force level of a task is given.
All slopes and levels are nonnegative, and the offsets of the sticking experts are bounded,
$0 \le b_j \le b_{\max}$.
Within the demonstrations of a single task, the normal force varies little with the
tangential load, so an unconstrained fit would place almost the entire force in the offset
and could not press harder on a heavier object.
The bound makes the slope carry the force, so that the law follows the measured load.

% ---------------------------------------------------------------------------
\subsubsection{Geometry-Gated Routing}
\label{sec:method:routing}

\moe{} selects one expert at every control step in two stages.
Geometric gates first exclude the experts whose condition in \cref{tab:experts} does not
hold, and a learned router then chooses among the remaining ones.
The gates enter the router as a mask $m_k$, whose entry is $0$ for an admissible expert
and $-\infty$ otherwise.
Since the conditions describe the contact rather than the task, the admissible experts
change within an episode.
During an insertion, for instance, the plug first slides over the surface while searching
for the hole, and the insertion expert becomes admissible only once the force aligns with
the tool axis.

The router $r_\omega$ scores the experts from the current image $I_k$ and the history
$\mathcal{H}_k \in \mathbb{R}^{8 \times 2}$ of the tangential force $F_t$ and the
end-effector speed over the last eight control steps, and the gated probabilities are
\begin{equation}
  p_k = \mathrm{softmax}\big(r_\omega(I_k, \mathcal{H}_k) + m_k\big).
  \label{eq:router}
\end{equation}
The router deliberately ignores the normal force and the direction of the force.
Both reflect the robot's own command, so a router that observed them could choose a
low-force expert merely because the robot currently presses lightly, which would keep the
force low.
For the same reason, the level of a sliding expert is chosen from the image alone, since
in a sliding contact the tangential force is caused by the normal force itself.
During execution, the expert with the highest probability is selected, and it changes only
when another one leads by a margin $\delta$ for several consecutive steps or when its gate
excludes it.

% ---------------------------------------------------------------------------
\subsubsection{Training and Execution}
\label{sec:method:training}

The only supervision is the normal force recorded in the demonstrations, and no frame is
labeled with its contact case or task.
The experts and the router are trained jointly as a single model on the contact frames of
all tasks.
The target of frame $k$ is the largest normal force over the next action chunk,
$F^\star_{n,k} = \max_{0 < i \le H_a} \bar{F}_{n,k+i}$, where $\bar{F}_n$ is the smoothed
normal force of the motion frame.
At the onset of contact, the experts thus learn the force about to be required rather than
the force built up so far.

The router makes a hard choice of one admissible expert $j_k$ per frame, trained
together with the experts through the straight-through Gumbel-softmax
estimator~\cite{DBLP:conf/iclr/JangGP17,DBLP:conf/iclr/MaddisonMT17}.
With $\hat{F}_{n,k} = f_{j_k}(k)$ the force predicted by the selected expert, we minimize
\begin{equation}
  \mathcal{L} = \mathbb{E}_k \big[ \big(\hat{F}_{n,k} - F^\star_{n,k}\big)^2
  + \lambda\, \pi_k \big],
  \label{eq:moe_loss}
\end{equation}
where $\lambda$ weights a level penalty with $\pi_k = 1$ if a constant-level expert, a pushing
or a sliding expert, is selected on a frame where a sticking expert is admissible, and
$\pi_k = 0$ otherwise.

Since only the selected expert receives its gradient, each expert is fitted to the frames
that it wins, and the router learns to prefer the admissible expert that fits better, so
the experts specialize.
The second term resolves an ambiguity of the data.
At a single block mass, a constant force and a sticking law fit the demonstrations of
pivoting almost equally well, but only the sticking law presses harder on a heavier block.
The penalty makes a constant-level expert win such a frame only if its squared error is lower
by more than $\lambda$.
Wiping, which truly slides, is still assigned to a sliding expert (Appendix~\ref{sec:laws}).

During execution, the hybrid controller follows the planned pose in every direction
except $z$, along which it applies the normal force while the end-effector is in contact
\begin{equation}
  F^{\mathrm{cmd}}_{n,k} = \mathrm{clip}\Big(
    w\,\big(c_j\, \psi_j + b_j\big) + (1 - w)\, \hat{F}_{n,k},\;
    0,\; F_{\max}\Big),
  \label{eq:command}
\end{equation}
where $j$ is the selected expert of~\eqref{eq:experts}, evaluated on the measured load,
$\hat{F}_{n,k}$ is the normal component of the wrench predicted by the global force
policy, which contributes the timing of the force within the chunk, $w \in [0, 1]$ weights
the two terms, and $F_{\max}$ is the largest normal force that may be commanded.
In free space, the motion frame is inactive, so no expert is selected and the planned
pose is tracked along $z$ as well.

\section{Experiments}
\label{sec:experiments}

We evaluate Force2Act in simulation and on a real robot to answer four questions:
\begin{itemize}
  \item \textbf{Q1:} How does Force2Act perform against contact-rich, visuomotor, and VLA
        baselines?
  \item \textbf{Q2:} Does Force2Act generalize to interactions that differ from the
        demonstrations?
  \item \textbf{Q3:} How do the individual components of the policy design contribute to
        performance?
  \item \textbf{Q4:} Does Force2Act transfer to a real robot?
\end{itemize}

\subsection{Experiment Setting}
\label{sec:exp:benchmark}

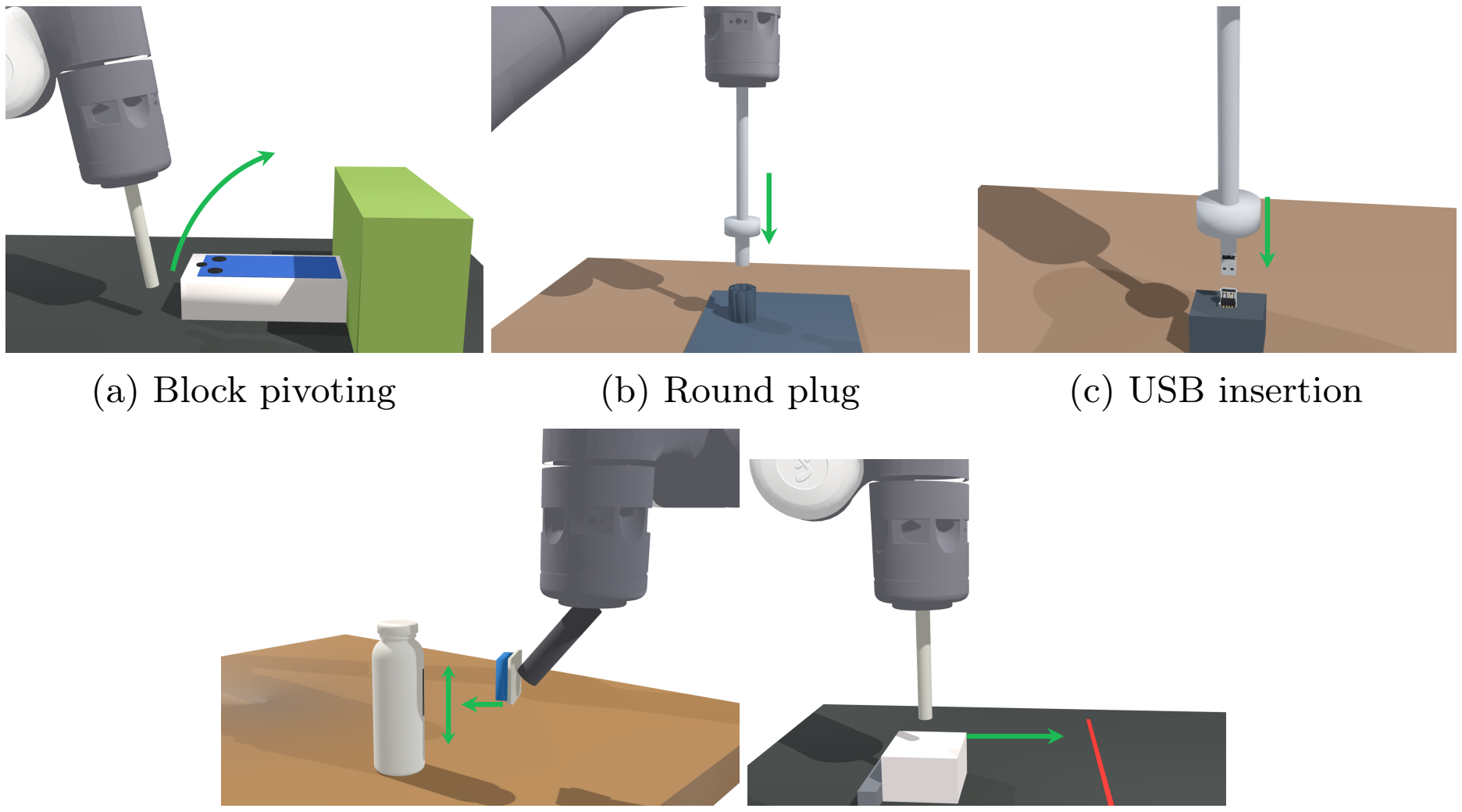
\begin{figure}[!htb]
  \centering
  \setlength{\tabcolsep}{1pt}
  % \taskimg{file}{overlay}: overlay coordinates are pixels of the 1600 x 1200 render
  % (origin at the top-left corner, y pointing down); rows 0 to 1160 are shown, so that the
  % wrist of the robot stays in view.
  \definecolor{cMot}{HTML}{1DB954}
  \newcommand{\taskimg}[2]{%
    \begin{tikzpicture}[baseline=(current bounding box.south),
        x=0.00016\linewidth, y=-0.00016\linewidth,
        mot/.style={-{Stealth[length=4pt, width=4pt]}, line width=1.2pt, draw=cMot},
        mot2/.style={{Stealth[length=3pt, width=3.5pt]}-{Stealth[length=3pt, width=3.5pt]},
                     line width=1.2pt, draw=cMot}]
      \useasboundingbox (0,0) rectangle (1600,1160);
      \clip (0,0) rectangle (1600,1160);
      \node[inner sep=0, anchor=north west] at (0,0)
        {\includegraphics[width=0.256\linewidth]{figs/benchmark/#1.png}};
      #2
    \end{tikzpicture}}
  % \taskimgz{file}{overlay}: same tile, zoomed 1.2x on the region (200,33)-(1533,1000)
  \newcommand{\taskimgz}[2]{%
    \begin{tikzpicture}[baseline=(current bounding box.south),
        x=0.000192\linewidth, y=-0.000192\linewidth,
        mot/.style={-{Stealth[length=4pt, width=4pt]}, line width=1.2pt, draw=cMot},
        mot2/.style={{Stealth[length=3pt, width=3.5pt]}-{Stealth[length=3pt, width=3.5pt]},
                     line width=1.2pt, draw=cMot}]
      \useasboundingbox (200,33) rectangle (1533,1000);
      \clip (200,33) rectangle (1533,1000);
      \node[inner sep=0, anchor=north west] at (0,0)
        {\includegraphics[width=0.3072\linewidth]{figs/benchmark/#1.png}};
      #2
    \end{tikzpicture}}
  \begin{tabular}{ccc}
    \taskimg{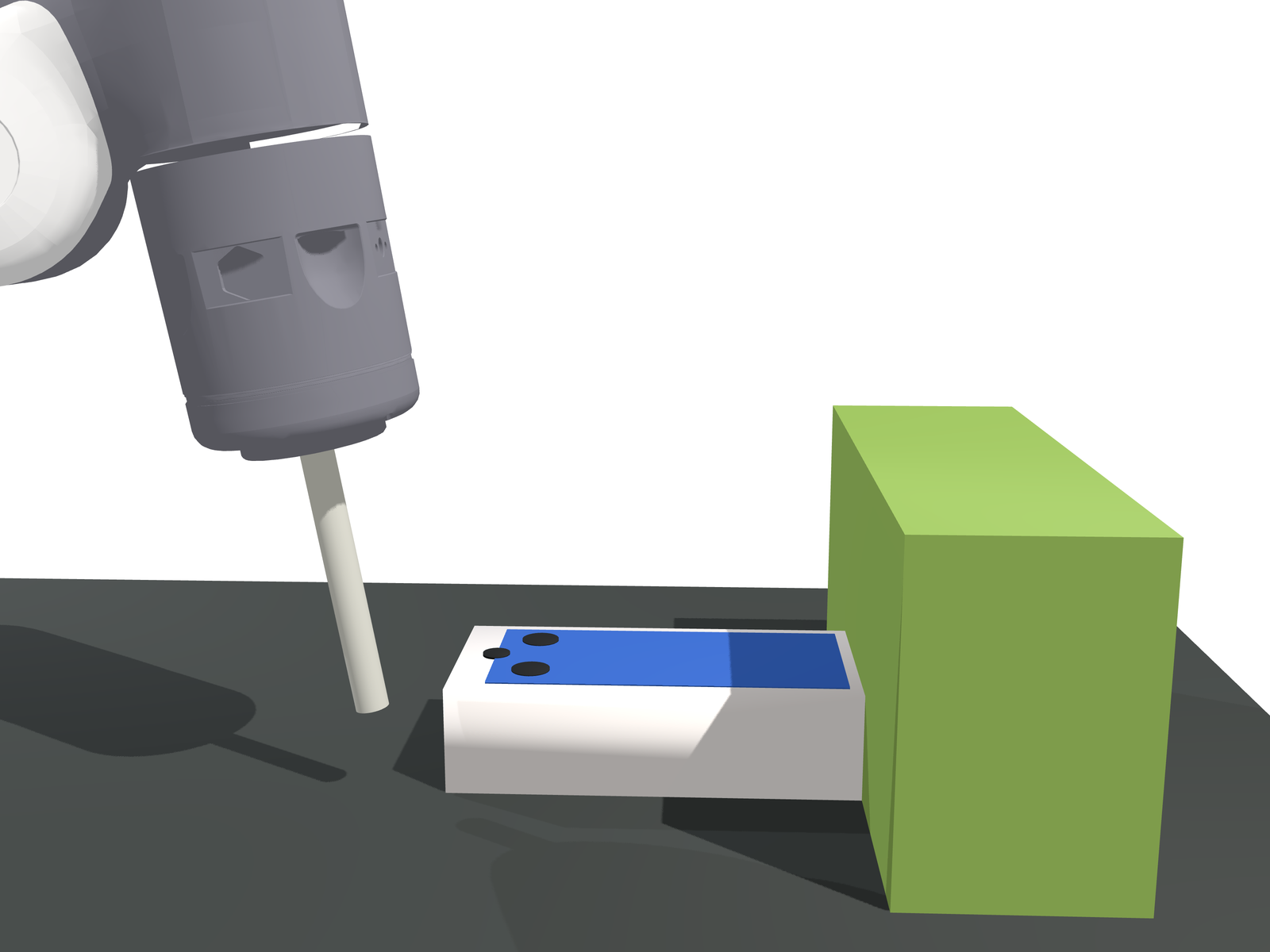}{\draw[mot] (562,888) arc[start angle=-168, end angle=-110, radius=540];} &
    \taskimg{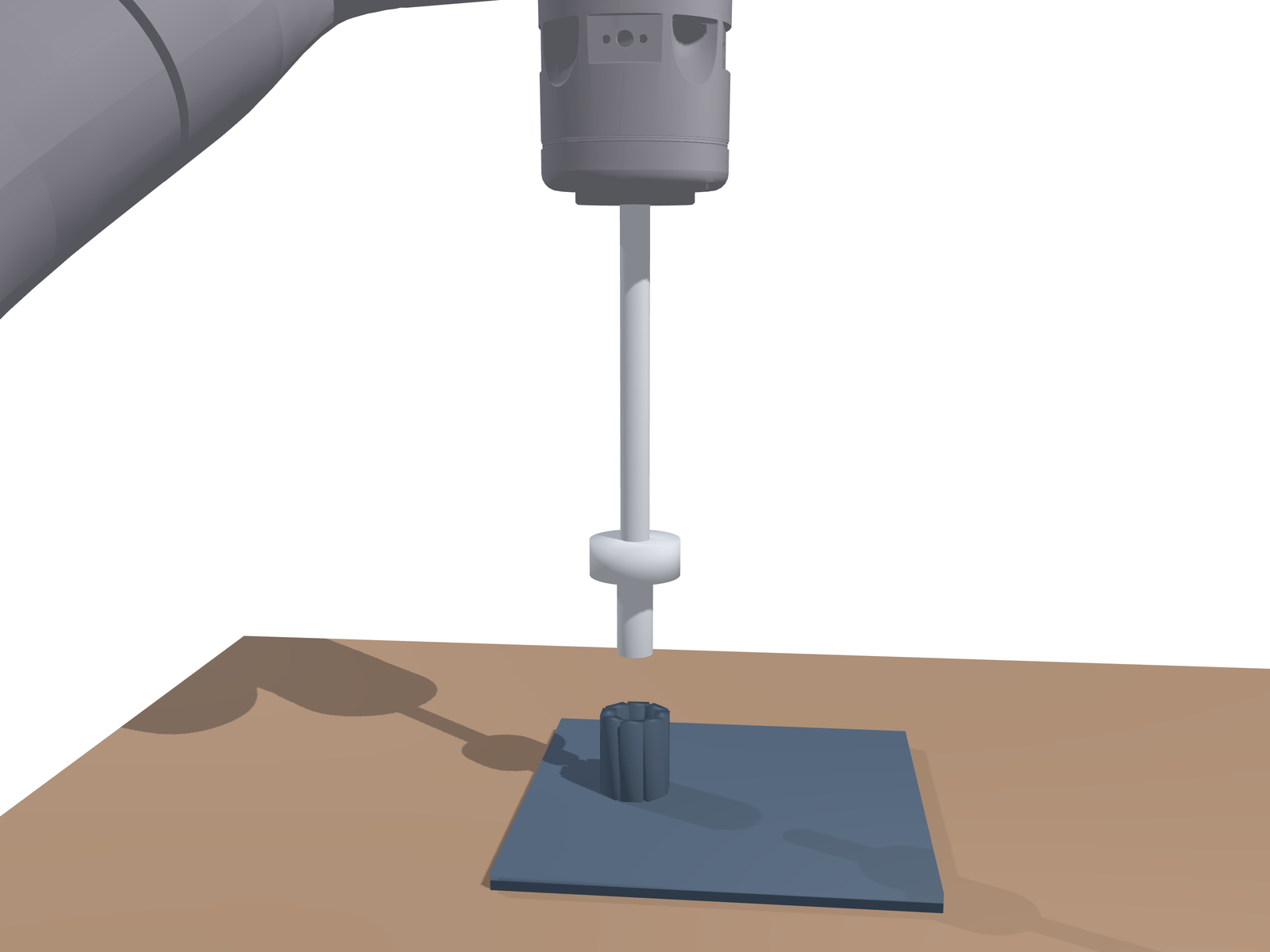}{\draw[mot] (930,560) -- (930,800);} &
    \taskimg{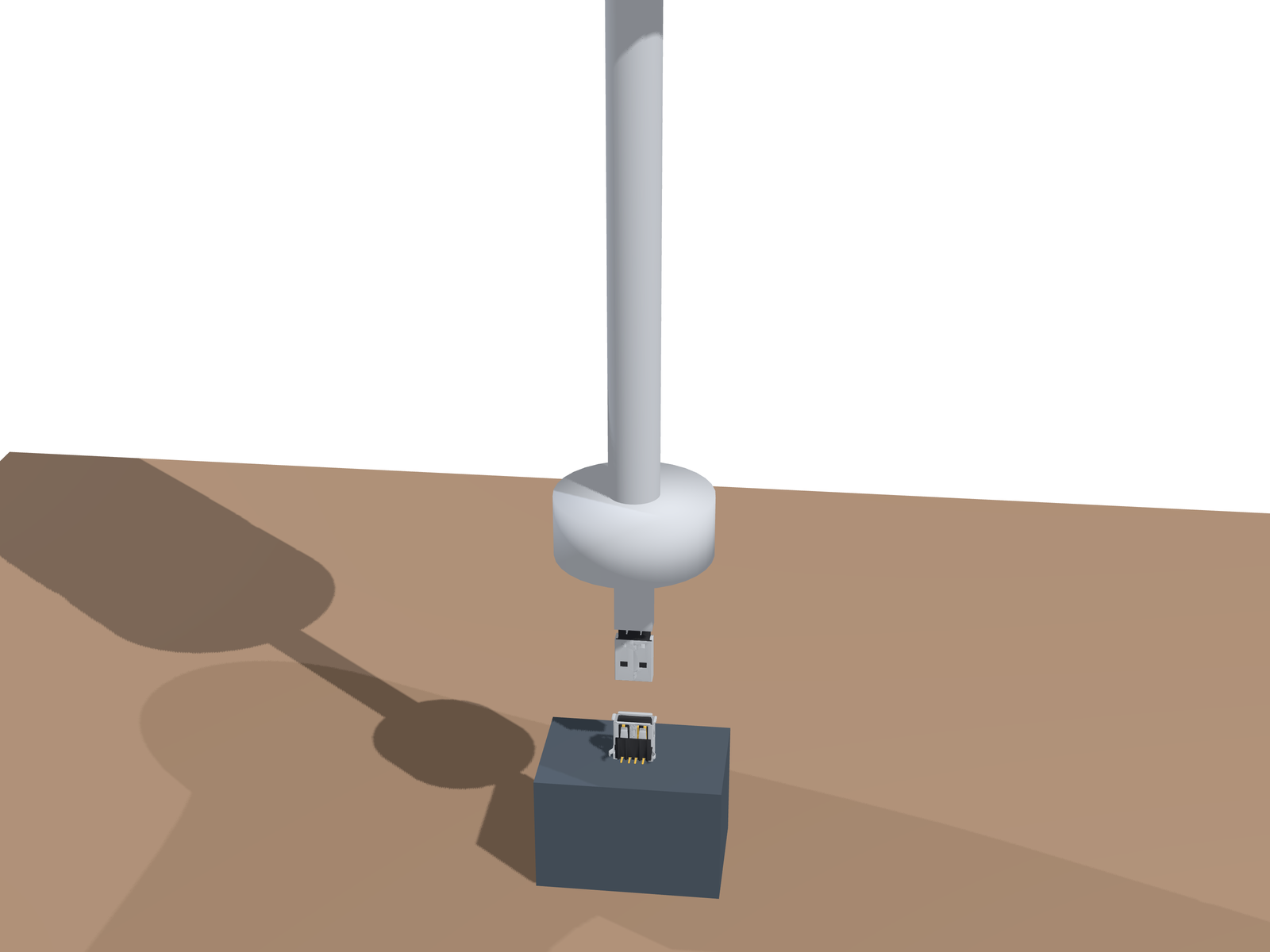}{\draw[mot] (970,640) -- (970,880);} \\
    \small (a) Block pivoting & \small (b) Round plug & \small (c) USB insertion \\[2pt]
  \end{tabular}
  \begin{tabular}{cc}
    \taskimgz{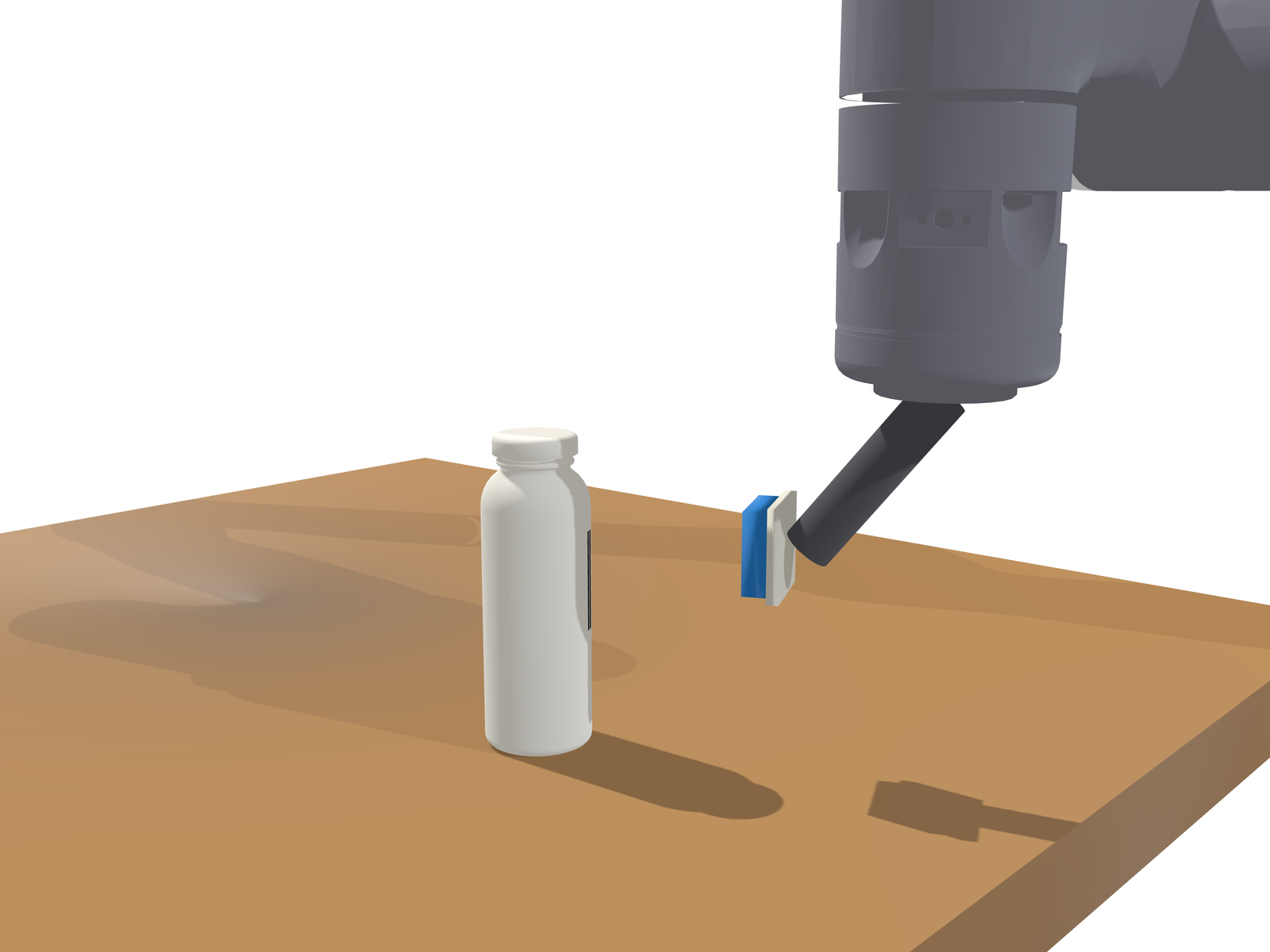}{\draw[mot] (925,741) -- (818,741);
                             \draw[mot2] (785,640) -- (785,845);} &
    \taskimg{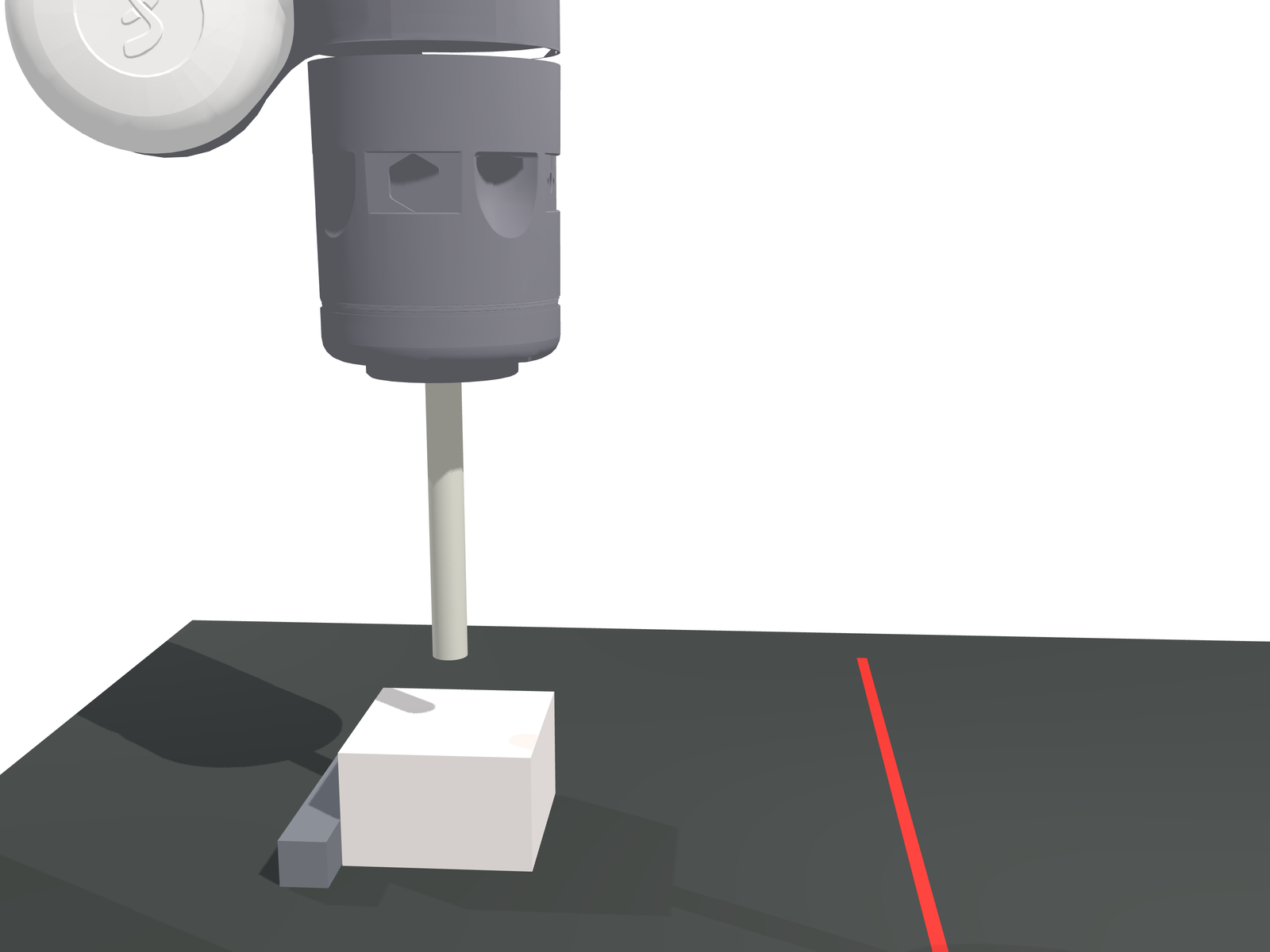}{\draw[mot] (735,930) -- (1060,930);} \\
    \small (d) Bottle wiping & \small (e) Press-push box
  \end{tabular}
  \vspace{-0.7em}
  \caption{Tasks of Force2Act-Bench, simulated in Genesis~\cite{genesis2024}.
    Arrows indicate the motion that solves each task.}
  \label{fig:benchmark}
  \vspace{-0.9em}
\end{figure}

\paragraph{Tasks and evaluation metrics.}
Force2Act-Bench consists of five contact-rich tasks simulated in Genesis~\cite{genesis2024}
(\cref{fig:benchmark}), each trained from teleoperated demonstrations with force feedback.
The tasks are designed so that reproducing the demonstrated motion alone does not suffice.
\emph{Block pivoting} pushes a block against a wall until it stands upright, over six block
masses from 0.4 to 0.9\,kg.
\emph{Round-plug insertion} inserts a plug into a hole at the demonstrated position or
displaced by 2\,mm.
\emph{USB insertion} inserts a USB connector into a socket with a 1\,mm tolerance.
\emph{Bottle wiping} removes an ink line from a free-standing bottle without pushing the bottle
away.
\emph{Press-push box} pushes a low-friction box uphill by pressing only on its top face,
with demonstrations on a flat support and evaluation on inclines of $0^\circ$, $2^\circ$, and
$4^\circ$.
The metric is the success rate over 305 trials with fixed seeds, identical for all policies,
and the success criteria are given in Appendix~\ref{sec:protocol}.

\newcommand{\nab}{\textcolor{gray}{--}}
\newcommand{\oursrow}{\rowcolor{cyan!8}}

\paragraph{Baselines.}
We compare against six policies from both families discussed in \cref{sec:related}.
Among visuomotor policies, ManiFlow~\cite{yan2025maniflow} uses no force,
RDP~\cite{xue2025rdp} and ImplicitRDP~\cite{chen2026implicitrdp} react to force through
slow--fast designs, and ACP~\cite{hou2025adaptive} predicts the stiffness of a compliance
controller from force feedback.
Among VLAs, ForceVLA~\cite{yu2025forcevla} fuses force into $\pi_0$, and
MolmoAct2~\cite{DBLP:journals/corr/abs-2605-02881} uses no force.
We also evaluate GPT-6.2-Astra~\cite{openai2026gpt62astra} zero-shot
to test how well a general-purpose model generalizes to these tasks.
All other policies are trained on the same demonstrations, and every policy is evaluated under
identical conditions, each baseline with the low-level controller it was designed for.
Every policy observes a single external RGB camera, without a wrist camera or depth, so
the contact cannot be resolved from vision and must be inferred from force.
We do not compare against Force Policy~\cite{fang2026forcepolicy}, which relies on depth
input, nor against ForceVLA2~\cite{li2026forcevla2} and CompVLA~\cite{kim2026compvla}, whose
code is not publicly available.

\subsection{Simulation Results}
\label{sec:exp:results}

\paragraph{Overall performance (Q1).}
In \cref{tab:main}, Force2Act achieves the highest overall success rate, 93.8\% against 77.0\%
for ACP, and is the best method on both tasks that test interaction generalization, block
pivoting and press-push box.
Apart from GPT-6.2-Astra, no VLA or position-controlled policy succeeds on block pivoting, and
they rarely succeed on bottle wiping, including ForceVLA, which observes force but does not
command it.
These policies fail on block pivoting even at the demonstrated mass, since the block must be
held against the wall by a sustained press within strict tolerances, whereas a position target
applies force only through its tracking error.
GPT-6.2-Astra succeeds on block pivoting in 30.7\% of trials and on bottle wiping in 58.0\%,
far below Force2Act on both.
USB insertion, whose 1\,mm tolerance is governed by lateral alignment rather than by the
normal force, is the only task on which baselines perform better.

\begin{table}[!htb]
  \centering
  \footnotesize
  \newcommand{\nt}[1]{{\scriptsize\textcolor{gray}{#1}}}
  \caption{\textbf{Success rates (\%) on Force2Act-Bench.}
    Block pivoting is averaged over six block masses, press-push box over three inclines, and
    round-plug insertion over the hole at its demonstrated position and displaced by 2\,mm.
    \emph{Avg.}\ is the success rate over all 305 trials, so each task is weighted by its number
    of trials, given below it.
    \textbf{Bold} marks the best result and \underline{underline} the second best.}
  \label{tab:main}
  \renewcommand{\arraystretch}{1.1}
  \setlength{\tabcolsep}{2.6pt}
  \begin{tabular}{l cccccc}
    \toprule
    Method & Pivot & Box & Wipe & Round & USB & Avg. \\[-1pt]
    \nt{Trials} & \nt{150} & \nt{30} & \nt{50} & \nt{50} & \nt{25} & \nt{305} \\
    \midrule
    \multicolumn{7}{l}{\emph{Without force}} \\
    \quad MolmoAct2~\cite{DBLP:journals/corr/abs-2605-02881} & 0.0 & 56.7 & 16.0 & 76.0 & 40.0 & 23.9 \\
    \quad ManiFlow~\cite{yan2025maniflow} & 0.0 & \underline{66.7} & 0.0 & 86.0 & \underline{76.0} & 26.9 \\
    \midrule
    \multicolumn{7}{l}{\emph{With force}} \\
    \quad ForceVLA~\cite{yu2025forcevla} & 0.0 & 33.3 & 4.0 & 88.0 & \underline{96.0} & 26.2 \\
    \quad RDP~\cite{xue2025rdp} & 0.0 & 0.0 & 0.0 & 86.0 & 0.0 & 14.1 \\
    \quad ImplicitRDP~\cite{chen2026implicitrdp} & 0.0 & 0.0 & 2.0 & 52.0 & 0.0 & 8.9 \\
    \quad ACP~\cite{hou2025adaptive} & \underline{64.0} & 50.0 & \textbf{100.0} & \textbf{100.0} & \underline{96.0} & \underline{77.0} \\
    \midrule
    GPT-6.2-Astra~\cite{openai2026gpt62astra} & 30.7 & \textbf{100.0} & 58.0 & 94.0 & \textbf{100.0} & 58.0 \\
    \midrule
    \oursrow
    \textbf{Force2Act (ours)} & \textbf{93.3} & \textbf{100.0} & \underline{98.0} & \underline{96.0} & 76.0 & \textbf{93.8} \\
    \bottomrule
  \end{tabular}
\end{table}

% Design ablation table of MoFE.

\paragraph{Interaction generalization (Q2).}
ACP succeeds with light blocks and on a flat support, but it degrades as the block becomes
heavier or the incline steeper, since the stiffness it learns reproduces the demonstrated
interaction rather than the load.
The other baselines fail on block pivoting at every mass, except GPT-6.2-Astra, whose success
falls from 40--44\% on lighter blocks to 16--24\% on heavier ones.
GPT-6.2-Astra is the only baseline that solves press-push box at every incline.
Force2Act keeps its success rate at every mass and incline, as its force follows the measured
load (\cref{tab:pivot,tab:box} in Appendix~\ref{sec:pertask}).

\subsection{Ablation Study}
\label{sec:exp:ablation}

\paragraph{Contribution of the two policies (Q3).}
\cref{tab:ablation} evaluates the two policies alone and combined through the weight $w$
in~\eqref{eq:command}.
The global force policy alone is weakest on block pivoting and press-push box, the two tasks
whose load changes at test time, since its plan is executed open loop within a chunk.
\moe{} alone solves the other tasks nearly as well as Force2Act but fails on block pivoting,
since it does not plan when and where the force should rise.
Their combination is the best overall and stable for every $w$ from 0.4 to 0.9, and we use
$w = 0.8$ throughout.
Only USB insertion, which is governed by lateral alignment rather than by the normal force,
does not benefit from the combination.
\cref{tab:failures} examines block pivoting, where the two policies differ most.
The global policy alone fails mostly because the block tips and falls back, as its plan
under-presses when the contact changes within a chunk, whereas \moe{} alone fails mostly by
stalling or without tipping the block.
Combined, Force2Act loses contact far less often, so \moe{} keeps the contact that the plan
relies on.

\begin{table}[!htb]
  \centering
  \footnotesize
  \caption{\textbf{Ablation of the policy weight and the force representation.}
    Success rate (\%) for the weight $w$ in~\eqref{eq:command}, from the global force policy
    alone ($w = 0$) to \moe{} alone ($w = 1$), with the global force policy trained on force
    relative to the measurement~\eqref{eq:scale} or on absolute force in newtons, over all 305
    trials in \emph{Avg.}}
  \label{tab:ablation}
  \renewcommand{\arraystretch}{1.1}
  \setlength{\tabcolsep}{3.2pt}
  \begin{tabular}{l cccccc}
    \toprule
    $w$ & Pivot. & Box & Wipe & Round & USB & Avg. \\
    \midrule
    \multicolumn{7}{l}{\emph{Relative force}} \\
    0 (global only) & 80.0 & 83.3 & 98.0 & 98.0 & 88.0 & 86.9 \\
    0.2 & 82.7 & 96.7 & \textbf{100.0} & \textbf{100.0} & 88.0 & 90.2 \\
    0.4 & 88.0 & \textbf{100.0} & 98.0 & \textbf{100.0} & 92.0 & 93.1 \\
    0.6 & 87.3 & \textbf{100.0} & \textbf{100.0} & \textbf{100.0} & 84.0 & 92.5 \\
    \oursrow
    0.8 (ours) & \textbf{93.3} & \textbf{100.0} & 98.0 & 96.0 & 76.0 & \textbf{93.8} \\
    0.9 & 90.0 & 96.7 & \textbf{100.0} & 98.0 & 72.0 & 92.1 \\
    1 (\moe{} only) & 38.0 & \textbf{100.0} & 94.0 & 98.0 & \textbf{96.0} & 67.9 \\
    \midrule
    \multicolumn{7}{l}{\emph{Absolute force}} \\
    0 (global only) & 29.3 & 46.7 & \textbf{100.0} & \textbf{100.0} & 60.0 & 56.7 \\
    0.8 & 78.7 & 83.3 & \textbf{100.0} & \textbf{100.0} & 92.0 & 87.2 \\
    \bottomrule
  \end{tabular}
\end{table}

\begin{table}[!htb]
  \centering
  \footnotesize
  \caption{\textbf{Failure modes on block pivoting.}
    150 trials per method over six masses (global only $w = 0$, \moe{} only $w = 1$).
    A failed trial ends as \emph{fell back} (the block drops over $20^\circ$ below its highest
    tilt), \emph{stalled} (timeout), or \emph{no tip} (tilt never above $10^\circ$).
    \emph{Contact loss} is $|F| < 1$\,N after first contact, as the share of trials and per trial.}
  \label{tab:failures}
  \renewcommand{\arraystretch}{1.1}
  \setlength{\tabcolsep}{2.2pt}
  \begin{tabular}{l cccc cc}
    \toprule
    & \multicolumn{4}{c}{Failed trials} & \multicolumn{2}{c}{Contact loss} \\
    \cmidrule(lr){2-5} \cmidrule(lr){6-7}
    Method & Total & Fell back & Stalled & No tip & Trials & Per trial \\
    \midrule
    ACP~\cite{hou2025adaptive} & 54 & 6 & 7 & 41 & 41\% & 1.00 \\
    Global only & 30 & 24 & 5 & 1 & 59\% & 1.97 \\
    \moe{} only & 93 & 18 & 52 & 23 & 67\% & 1.01 \\
    \oursrow
    Force2Act (ours) & \textbf{10} & 9 & 1 & 0 & \textbf{22\%} & \textbf{0.29} \\
    \bottomrule
  \end{tabular}
\end{table}

\paragraph{Relative force (Q3).}
To isolate the relative force~\eqref{eq:scale}, we train the global force policy on absolute
force in newtons, with all else unchanged (\cref{tab:ablation}).
A policy that regresses absolute force reproduces the magnitudes of its demonstrations, so its
plan cannot press harder for a load it has not seen.
Alone, it therefore still solves bottle wiping and round-plug insertion, whose required force
matches the demonstrations, but fails on block pivoting and press-push box, whose load changes
with the mass and the incline.
Adding \moe{}, which raises the force with the measured load between plans, recovers most of
these failures.
The plan that \moe{} corrects, however, still starts from the demonstrated magnitude, so the
force grows less with the load than in Force2Act, as the heaviest blocks show
(\cref{tab:absforce} in Appendix~\ref{sec:pertask}).
The relative force lets the plan scale with the load, and \moe{} keeps the force scaled
between plans, so both are needed.

\paragraph{Design of \moe{} (Q3).}
\cref{tab:design} starts from a law without the structure of \moe{} and then removes one
component of \moe{} at a time.
Each component handles a particular kind of contact, so each variant loses different tasks of
the benchmark.
The MLP law observes the image and the full force history, including the force direction, but
instead of selecting a force expert it predicts a free slope and offset,
$F_n = \alpha_k |F_t| + \beta_k$, at every step.
Since both terms are free and its inputs include the force that the robot itself commands, it
can fit the demonstrations by reproducing the current force rather than by following the load.
It therefore solves bottle wiping and round-plug insertion, whose force matches the
demonstrations, but fails more often on block pivoting and press-push box, whose load changes
at test time, although it is trained for each task separately.

\begin{table}[!htb]
  \centering
  \footnotesize
  \caption{\textbf{Design ablations of \moe{}.}
    Success rate (\%) at $w = 0.8$.
    The MLP law replaces \moe{} by a free slope and offset predicted at every step and is
    trained for each task separately.
    Each further row removes one component of \moe{} from Force2Act.}
  \label{tab:design}
  \renewcommand{\arraystretch}{1.1}
  \setlength{\tabcolsep}{2.4pt}
  \begin{tabular}{l cccccc}
    \toprule
    Variant & Pivot & Box & Wipe & Round & USB & Avg. \\
    \midrule
    MLP law (no \moe{}) & 81.3 & 66.7 & \textbf{100.0} & \textbf{100.0} & 84.0 & 86.2 \\
    \midrule
    \multicolumn{7}{l}{\emph{Force2Act with one component removed}} \\
    \quad No geometric gates & 31.3 & \textbf{100.0} & 2.0 & 96.0 & 88.0 & 48.5 \\
    \quad Router sees direction & 58.7 & 96.7 & \textbf{100.0} & \textbf{100.0} & 88.0 & 78.4 \\
    \quad One expert per case & 83.3 & 96.7 & 22.0 & \textbf{100.0} & \textbf{92.0} & 78.0 \\
    \quad No level penalty & 67.3 & 76.7 & \textbf{100.0} & 98.0 & 84.0 & 80.0 \\
    \quad No offset bound & 90.0 & 96.7 & 98.0 & \textbf{100.0} & 80.0 & 92.8 \\
    \midrule
    \oursrow
    Force2Act (ours) & \textbf{93.3} & \textbf{100.0} & 98.0 & 96.0 & 76.0 & \textbf{93.8} \\
    \bottomrule
  \end{tabular}
\end{table}

A force expert makes the force follow the contact only if it is the right expert, which is the
role of the gates and the router.
The gates rule out the experts whose geometry contradicts the contact.
Without them, the router must tell all contact cases apart from the image and the force
history alone, and it fails to do so.
The force on block pivoting then stays flat at every mass, and bottle wiping almost never
succeeds, so this variant falls the furthest.
When the router also observes the force direction, which reflects the robot's own
command (\cref{sec:method:routing}), it can select a low-force expert merely
because the robot currently presses lightly, which keeps the force low.
This mainly affects block pivoting, where the heavier blocks lose contact in most trials.

The selected expert must also have the right constants, which is why each contact case has
several experts.
With one expert per case, a single sliding level forces the same press on every sliding
contact, whereas bottle wiping removes the ink only under a sustained press and pushes the
bottle away when pressed too hard, so wiping fails in most trials.
A single sticking law likewise forces one ratio of normal to tangential force on press-push box
and block pivoting alike, which under-presses the heaviest block.
Finally, the level penalty $\lambda$ in~\eqref{eq:moe_loss} keeps a constant-level expert from being
selected when another expert explains the frame almost as well.
Without it, the laws fit the held-out demonstrations of block pivoting more closely, yet a
constant level cannot grow with the load, so both sticking tasks, block pivoting and press-push
box, fail more often as the mass or the incline increases.
The offset bound $b_{\max}$ keeps the offset of a sticking expert from acting as a preload.
Without it, a larger part of the force is fixed rather than proportional to the load, which
suffices on most of the benchmark but under-presses the heaviest block.

\subsection{Real-World Results}
\label{sec:exp:hardware}

\begin{table}[!htb]
  \centering
  \footnotesize
  \caption{\textbf{Real-world results.}
    Successful trials per condition on the real robot, and the average success rate (\%) over
    all 120 trials.
    The demonstrations use the 0.116\,kg block and the flat table, whereas the 0.418\,kg block
    and the incline change the interaction.}
  \label{tab:hardware}
  \renewcommand{\arraystretch}{1.1}
  \setlength{\tabcolsep}{2.2pt}
  \begin{tabular}{ll ccc >{\columncolor{cyan!8}}c}
    \toprule
    Task & Condition & ACP & ForceVLA & ManiFlow & Force2Act \\
    \midrule
    \multirow{2}{*}{\makecell[l]{Block\\pivoting}} & 0.116\,kg & 18/20 & 0/20 & 3/20 & \textbf{19/20} \\
      & 0.418\,kg & 3/20 & 0/20 & 0/20 & \textbf{20/20} \\
    \cmidrule{1-6}
    \multirow{2}{*}{\makecell[l]{Press-push\\box}} & Flat table & 0/20 & 14/20 & 10/20 & \textbf{18/20} \\
      & $10^\circ$ incline & 1/20 & 0/20 & 17/20 & \textbf{20/20} \\
    \cmidrule{1-6}
    Press-push T & Demo setup & 8/20 & 0/20 & 3/20 & \textbf{20/20} \\
    \cmidrule{1-6}
    Plug insertion & Demo setup & 2/20 & 13/20 & 0/20 & \textbf{16/20} \\
    \midrule
    \multicolumn{2}{l}{Average success rate (\%)} & 26.7 & 22.5 & 27.5 & \textbf{94.2} \\
    \bottomrule
  \end{tabular}
\end{table}

\paragraph{Real-world performance (Q4).}
We evaluate Force2Act, ACP, ForceVLA, and ManiFlow on a real Flexiv Enlight~L arm, with 20
trials per condition (\cref{tab:hardware}).
Each policy is trained on 50 demonstrations of block pivoting and 25 of each other task, and
the object positions are kept as in the demonstrations.
In \emph{press-push T}, the tool presses on a T-shaped block and carries it across the table.
Block pivoting and press-push box are also tested out of distribution (OOD), with a 0.418\,kg block
instead of the demonstrated 0.116\,kg one and with the box on a $10^\circ$ incline that rises along
the push direction.
Force2Act reaches an average success rate of 94.2\%, against at most 27.5\% for the
baselines, and is the best in every condition.
ACP pivots the demonstrated block but rarely the heavier one, whereas Force2Act succeeds with
both, as its force follows the load.
On press-push box, ACP presses with about the demonstrated force, but its tool tilts and
slides off the top face, whereas ForceVLA in about a third of its trials never presses and
pushes the box from behind instead.
ManiFlow succeeds on the incline only by pressing far harder than in the demonstrations, as
its position target stays below the raised top face, whereas Force2Act keeps a light press on
both supports.
On plug insertion, the successful trials of ForceVLA mostly stop short of the demonstrated
depth while pushing hard against the socket, and ManiFlow never touches the socket.

\begin{figure*}[!htb]
  \centering
  \input{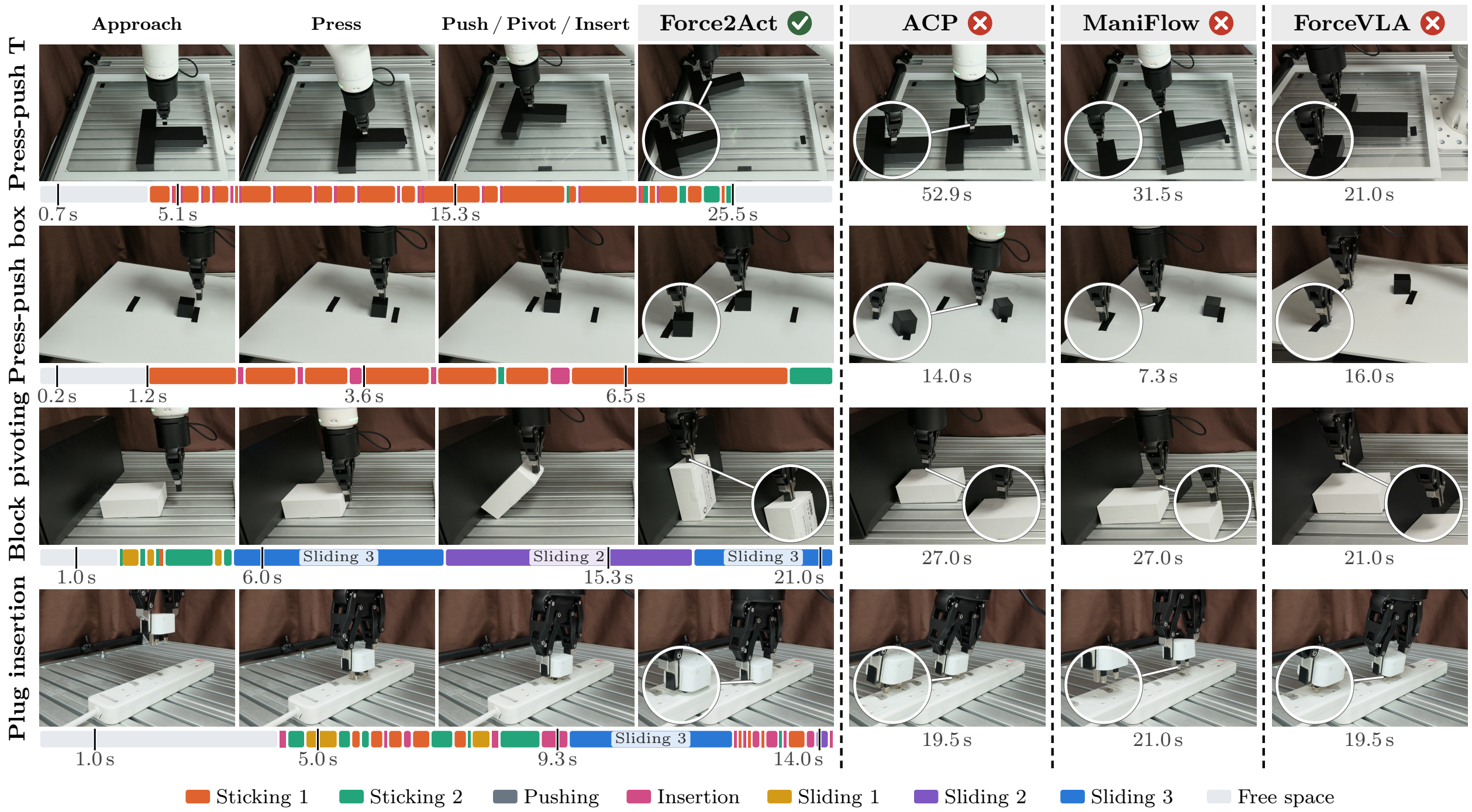}
  \vspace{-1em}
  \caption{\textbf{Real-world rollouts on four tasks.}
    The first three columns show a Force2Act rollout, and the last four show the outcome
    of each policy, with the contact enlarged.
    Force2Act completes all four tasks, whereas ACP gets stuck or loses the object, ManiFlow
    keeps moving without applying enough force, and ForceVLA misses the contact or stops on
    the face of the socket.
    Below each Force2Act rollout, the expert selected over time, with a tick at the time of
    each frame in the video.}
  \label{fig:real}
  \vspace{-1.2em}
\end{figure*}

\FloatBarrier
\section{Conclusion}
\label{sec:conclusion}
We present Force2Act, a force-aware visuomotor policy that combines a global force policy,
which plans the motion along with a relative interaction force, with the Mixture of Force
Experts, which regulates the force at every control step.
We evaluate it in simulation on Force2Act-Bench and on a real robot.
Force2Act achieves an overall success rate of 93.8\% in simulation and 94.2\% on the real
robot, outperforming the baselines, and keeps its success rate with heavier objects and on inclined supports beyond its demonstrations.
Ablations show that both policies, the relative force, and each component of the Mixture of
Force Experts are needed.
One limitation of this work is that Force2Act regulates the force along a single direction,
the contact normal, so tasks that require compliance in multiple directions or rotational
compliance remain future work.
Furthermore, all contacts in our tasks are rigid, so soft or deformable contacts are not
considered.

%%%%%%%%% REFERENCES %%%%%%%%%
\bibliographystyle{abbrvnat}
\bibliography{main}

%%%%%%%%% APPENDIX %%%%%%%%%
\clearpage
\beginappendix
\noindent All results in this appendix are obtained in simulation on
Force2Act-Bench, except where a sentence explicitly refers to the real robot.
The real-robot results are reported in \cref{sec:exp:hardware}.

\section{Implementation Details}
\label{sec:implementation}

\subsection{Demonstrations and Evaluation}
\label{sec:protocol}

\paragraph{Demonstrations.}
The demonstrations are collected with a custom force-feedback teleoperation system, rather
than the Flexiv TDK teleoperation kit, in which a real Flexiv Rizon~4 arm serves as a compliant
teleoperation device.
The operator guides the arm by hand, and the simulated robot follows its end-effector pose.
The contact force measured in simulation is rendered back on the arm, so that the operator
feels the interaction, which makes the contact-rich tasks easy to perform.
We record 50 demonstrations of block pivoting, round-plug insertion, and USB insertion, 65 of
bottle wiping, and 75 of press-push box (\cref{tab:protocol}), and every policy is trained on
all of them, without a validation split.

\paragraph{Task design.}
The tolerances of the tasks are deliberately strict, so that a small error in motion or force
loses the contact or fails the task, and a policy that succeeds in simulation is therefore
more likely to be robust on a real robot.
In block pivoting, lighter blocks are excluded, since they can be pivoted by reproducing the
demonstrated force and therefore do not test force regulation.
In bottle wiping, only a sustained press removes the ink, while pressing too hard pushes the
bottle away.
\cref{tab:protocol} lists the trials of each task.

\paragraph{Success criteria.}
Block pivoting succeeds when the block stands within $2^\circ$ of vertical, round-plug
insertion when the plug is at least 30\,mm deep within 2\,mm of the hole axis, and USB
insertion when the connector is at least 9.4\,mm deep within 1\,mm of the socket axis.
Bottle wiping succeeds when at least 90\% of the ink line is removed while the bottle moves by
at most 15\,mm, and press-push box when the box has been pushed 155\,mm uphill.
On the real robot, press-push box pushes the box 11.9\,cm, and plug insertion succeeds when
the plug reaches within about 10\,mm of the demonstrated depth.

\begin{table}[!htb]
  \centering
  \footnotesize
  \caption{\textbf{Demonstrations and evaluation protocol.}
    Demonstrations and evaluation trials of each task, with fixed seeds identical for all
    policies.
    The ink lines of bottle wiping are shifted by up to $\pm$5\,mm.}
  \label{tab:protocol}
  \renewcommand{\arraystretch}{1.1}
  \setlength{\tabcolsep}{3.5pt}
  \begin{tabular}{l r l r}
    \toprule
    Task & Demos & Conditions & Trials \\
    \midrule
    Block pivoting & 50 & 6 masses, 25 each & 150 \\
    Press-push box & 75 & 3 inclines, 10 each & 30 \\
    Round-plug insertion & 50 & 2 hole positions, 25 each & 50 \\
    Bottle wiping & 65 & 50 ink lines, $\pm$5\,mm shift & 50 \\
    USB insertion & 50 & Demo socket, 25 & 25 \\
    \midrule
    Total & 290 & & 305 \\
    \bottomrule
  \end{tabular}
\end{table}

\paragraph{Training of the baselines.}
ManiFlow, RDP, ImplicitRDP, and ACP are trained for 300 epochs, whereas the global
force policy of Force2Act is trained for 60 and its reactive compliance policy for 10.
ForceVLA is fully fine-tuned from its pretrained weights, and MolmoAct2 is fine-tuned with
LoRA~\cite{hu2022lora} from its DROID~\cite{khazatsky2024droid} checkpoint, both for
15 epochs.
Force2Act is trained on an NVIDIA H200 GPU.

\subsection{Global Force Policy}

\cref{tab:hyper} lists the hyperparameters of both policies.

\begin{table*}[!tb]
  \centering
  \footnotesize
  \caption{\textbf{Hyperparameters of Force2Act.}}
  \label{tab:hyper}
  \renewcommand{\arraystretch}{1.1}
  \setlength{\tabcolsep}{3pt}
    \begin{tabular}{l l}
      \toprule
      \multicolumn{2}{l}{\textbf{Global force policy}} \\
      \midrule
      Control rate & 15\,Hz \\
      Observed images and poses & last 2 \\
      Observed wrenches & last 8 (0.53\,s) \\
      Action chunk $H$ / executed $H_a$ & 16 / 8 \\
      Relative-force scale $s_{\min}$ & 1\,N \\
      Image encoder & CLIP ViT-B/16, fine-tuned \\
      Instruction encoder & T5-small, frozen \\
      Context token width & 768 \\
      Action head & 12 blocks, width 768, 12 heads \\
      Integration & 10 Euler steps \\
      Epochs / batch size & 60 / 128 \\
      Optimizer & AdamW, decay $10^{-3}$ \\
      Peak learning rate & $1.4 \times 10^{-4}$ \\
      Schedule & 200 warm-up steps, cosine \\
      \bottomrule
    \end{tabular}\hfill
    \begin{tabular}{l l}
      \toprule
      \multicolumn{2}{l}{\textbf{Reactive compliance policy (\moe{})}} \\
      \midrule
      Router backbone & ResNet-18, $224 \times 224$ \\
      Force and speed MLP & width 128, last 8 steps \\
      Hidden layer & width 256 \\
      Offset bound $b_{\max}$ & 2\,N \\
      Training data & 30{,}968 contact frames \\
      Epochs / batch size & 10 / 256 \\
      Router optimizer & AdamW, $2 \times 10^{-4}$, decay 0.01 \\
      Constants optimizer & AdamW, $2 \times 10^{-2}$, no decay \\
      Gumbel temperature & 1 to 0.1, geometric \\
      Level penalty $\lambda$ & 20\,N$^2$ \\
      Router force scaling & $[0.5, 2]$ per sample \\
      Router feature jitter & $\pm 1\%$ \\
      Switching margin $\delta$ & 0.2 for 3 steps \\
      Weight $w$ / limit $F_{\max}$ & 0.8 / 60\,N \\
      \bottomrule
    \end{tabular}
\end{table*}

\paragraph{Encoders.}
Images are encoded by a CLIP ViT-B/16~\cite{radford2021clip}, and the instruction by a
frozen T5-small model.
The CLIP encoder is fine-tuned end-to-end and contributes the patch and class tokens of
each image.
Each wrench sample is embedded by a two-layer MLP and each pose by a linear layer, each to
its own token.

\paragraph{Action head.}
The action head is a diffusion transformer whose blocks apply self-attention over the action
chunk, cross-attention to the context tokens, and a feed-forward layer, each modulated by the
flow time through adaptive layer normalization.
Given a normalized action chunk $A$, Gaussian noise $\epsilon \sim \mathcal{N}(0, I)$, and a
flow time $\tau \in [0, 1]$ drawn from the shifted Beta distribution of
$\pi_0$~\cite{black2024pi0}, we form $A_\tau = (1-\tau)\,\epsilon + \tau A$ and minimize
\begin{equation}
  \Lfm(\phi, \theta) =
  \mathbb{E}_{\tau, \epsilon}\,
  \big\| v_\theta\big(A_\tau, \tau \mid E_\phi(o_t)\big) - (A - \epsilon) \big\|_2^2 ,
  \label{eq:fm}
\end{equation}
where $o_t = (\mathcal{I}_t, \mathcal{P}_t, \mathcal{W}_t / s_t, \ell)$.
The policy is trained separately for each task, with linear warm-up and cosine decay of the
learning rate.

\subsection{Reactive Compliance Policy}

\paragraph{Motion frame and contact features.}
A frame counts as a contact frame when the measured force at the next step is at least 1\,N.
The tangential force $F_t$ is the component of the measured force along the measured
velocity, and it is set to zero when the end-effector moves slower than 5\,mm/s or when the
force lies within $15^\circ$ of the motion, in which case the entire force counts as the
normal force.
These thresholds are set from the noise of the wrench and of the velocity on the
real robot, so that noise neither creates a contact nor tilts the motion frame there.
The force and the velocity in simulation are nearly noise-free, but the same thresholds are
used, so that the reactive compliance policy is identical in simulation and on the real
robot.

\paragraph{Gates.}
\cref{tab:gates} lists the condition under which each expert is admissible.

\begin{table}[!htb]
  \centering
  \footnotesize
  \caption{\textbf{Gates of the experts.}
    $\hat{n}$ is the normal of the surface the object rests on.
    All our supports are horizontal or tilted by at most $10^\circ$, so we use the upward
    vertical as $\hat{n}$.
    $F_t > 0$ means that the tool is moving ($\geq 5$\,mm/s) and that the force is more than
    $15^\circ$ from the motion.}
  \label{tab:gates}
  \renewcommand{\arraystretch}{1.1}
  \begin{tabular}{@{}l p{0.5\textwidth}@{}}
    \toprule
    Expert & Admissible when \\
    \midrule
    Pushing & the force has been along the motion while moving in at least four of the last
    eight steps \\
    Sticking 1 & $F_t > 0$ and the reaction on the end-effector is within $45^\circ$ of
    $\hat{n}$ (pressing onto the object) \\
    Sticking 2 & $F_t > 0$ and the reaction is more than $45^\circ$ from $\hat{n}$ (pressing
    against a side face) \\
    Insertion & a force of at least 1\,N lies within $50^\circ$ of the tool axis and, while
    moving, the motion is within $30^\circ$ of it \\
    Sliding & always \\
    \bottomrule
  \end{tabular}
\end{table}

\paragraph{Router and experts.}
The router combines a fine-tuned ImageNet-pretrained ResNet-18 with an MLP over the
tangential force magnitude and the speed of the recent steps, followed by a hidden layer
that outputs the scores of the five contact cases.
The scores of the three sliding levels are predicted from the image features alone.
The last layers are initialized to zero, so that all admissible experts start equally
likely.
Slopes and levels are parameterized through a softplus and the offsets of the sticking experts
through a sigmoid scaled to $[0, b_{\max}]$.
Both slopes are initialized to the ratio of the median target to the median tangential
force over all frames, the offsets to 1\,N, the sliding levels to the 20th, 50th, and 80th
percentiles of the target, and the pushing expert to its median.

\paragraph{Training of the experts.}
The model is trained on the contact frames of the demonstrations of all five tasks.
The router and the constants of the experts use separate optimizer groups.
The Gumbel-softmax estimator samples a single expert from the gated scores $r_\omega + m_k$
in the forward pass and passes gradients to the router through the softmax in the backward
pass.
The target uses a centred five-step moving average of the normal force, padded with the edge
values at the boundaries of each episode, and its look-ahead of $H_a = 8$ steps stops at the
last frame of the episode.
To keep the routing independent of the magnitude of the load, the tangential force observed
by the router is scaled by a random factor per sample, and the router features are perturbed
by a small jitter.
With $\lambda = 60$\,N$^2$ instead of 20\,N$^2$, the penalty would assign bottle wiping to the
second sticking expert, since it would exceed the difference of about 45\,N$^2$ between the
squared errors of this expert and of a sliding level on wiping.

\paragraph{Execution.}
The held expert changes only when another leads by $\delta$ in probability for three
consecutive steps, which avoids switching on sensor noise, and immediately when its gate
excludes it.

% Per-task results. Move to the supplementary material before submission.
\section{Per-Task Results in Simulation}
\label{sec:pertask}

\begin{table}[!htb]
  \footnotesize
  \renewcommand{\arraystretch}{1.1}
  \centering
  \caption{\textbf{Block pivoting per block mass.}
    Success rate (\%) for each mass (kg), with the average over all six.
    Force2Act, $w = 0$ executes the global force policy alone.}
  \label{tab:pivot}
  \begin{tabular}{l cccccc c}
    \toprule
    Method & 0.4 & 0.5 & 0.6 & 0.7 & 0.8 & 0.9 & Avg. \\
    \midrule
    MolmoAct2~\cite{DBLP:journals/corr/abs-2605-02881} & 0 & 0 & 0 & 0 & 0 & 0 & 0.0 \\
    ManiFlow~\cite{yan2025maniflow} & 0 & 0 & 0 & 0 & 0 & 0 & 0.0 \\
    ForceVLA~\cite{yu2025forcevla} & 0 & 0 & 0 & 0 & 0 & 0 & 0.0 \\
    RDP~\cite{xue2025rdp} & 0 & 0 & 0 & 0 & 0 & 0 & 0.0 \\
    ImplicitRDP~\cite{chen2026implicitrdp} & 0 & 0 & 0 & 0 & 0 & 0 & 0.0 \\
    ACP~\cite{hou2025adaptive} & \underline{96} & \textbf{96} & \textbf{96} & \underline{76} & \underline{20} & 0 & \underline{64.0} \\
    GPT-6.2-Astra~\cite{openai2026gpt62astra} & 40 & 44 & 36 & 24 & 16 & \underline{24} & 30.7 \\
    Force2Act, $w = 0$ & 76 & 76 & 80 & 88 & 80 & 80 & 80.0 \\
    \oursrow
    \textbf{Force2Act (ours)} & \textbf{100} & \textbf{96} & \underline{92} & \textbf{92} & \textbf{88} & \textbf{92} & \textbf{93.3} \\
    \bottomrule
  \end{tabular}
\end{table}

\begin{table}[!htb]
  \footnotesize
  \renewcommand{\arraystretch}{1.1}
  \setlength{\tabcolsep}{3pt}
  \begin{minipage}[t]{0.52\textwidth}
    \centering
    \caption{\textbf{Press-push box per incline.}
      Success rate (\%) over 10 trials per incline. The demonstrations are recorded on a
      flat support.}
    \label{tab:box}
    \begin{tabular}{l ccc c}
      \toprule
      Method & $0^\circ$ & $2^\circ$ & $4^\circ$ & Avg. \\
      \midrule
      MolmoAct2~\cite{DBLP:journals/corr/abs-2605-02881} & \underline{10.0} & \underline{70.0} & \underline{90.0} & 56.7 \\
      ManiFlow~\cite{yan2025maniflow} & 0.0 & \textbf{100.0} & \textbf{100.0} & \underline{66.7} \\
      ForceVLA~\cite{yu2025forcevla} & 0.0 & 0.0 & \textbf{100.0} & 33.3 \\
      RDP~\cite{xue2025rdp} & 0.0 & 0.0 & 0.0 & 0.0 \\
      ImplicitRDP~\cite{chen2026implicitrdp} & 0.0 & 0.0 & 0.0 & 0.0 \\
      ACP~\cite{hou2025adaptive} & \textbf{100.0} & 40.0 & 10.0 & 50.0 \\
      GPT-6.2-Astra~\cite{openai2026gpt62astra} & \textbf{100.0} & \textbf{100.0} & \textbf{100.0} & \textbf{100.0} \\
      \oursrow
      \textbf{Force2Act (ours)} & \textbf{100.0} & \textbf{100.0} & \textbf{100.0} & \textbf{100.0} \\
      \bottomrule
    \end{tabular}
  \end{minipage}\hfill
  \begin{minipage}[t]{0.46\textwidth}
    \centering
    \caption{\textbf{Round-plug insertion.}
      Success rate (\%) with the hole at its demonstrated position (\emph{Demo}) and displaced
      by 2\,mm.}
    \label{tab:round}
    \begin{tabular}{l cc c}
      \toprule
      Method & Demo & $+$2\,mm & Avg. \\
      \midrule
      MolmoAct2~\cite{DBLP:journals/corr/abs-2605-02881} & 72.0 & 80.0 & 76.0 \\
      ManiFlow~\cite{yan2025maniflow} & \underline{96.0} & 76.0 & 86.0 \\
      ForceVLA~\cite{yu2025forcevla} & \textbf{100.0} & 76.0 & 88.0 \\
      RDP~\cite{xue2025rdp} & \underline{96.0} & 76.0 & 86.0 \\
      ImplicitRDP~\cite{chen2026implicitrdp} & 76.0 & 28.0 & 52.0 \\
      ACP~\cite{hou2025adaptive} & \textbf{100.0} & \textbf{100.0} & \textbf{100.0} \\
      GPT-6.2-Astra~\cite{openai2026gpt62astra} & \textbf{100.0} & 88.0 & 94.0 \\
      \oursrow
      \textbf{Force2Act (ours)} & \textbf{100.0} & \underline{92.0} & \underline{96.0} \\
      \bottomrule
    \end{tabular}
  \end{minipage}
\end{table}

\begin{table}[!htb]
  \footnotesize
  \renewcommand{\arraystretch}{1.1}
  \centering
  \caption{\textbf{Block pivoting per block mass for each weight $w$.}
    Success rate (\%) of Force2Act over 25 trials per mass (kg), with the average over all
    six, for the weight $w$ of the selected law in~\eqref{eq:command}.}
  \label{tab:pivot_w}
  \begin{tabular}{l cccccc c}
    \toprule
    $w$ & 0.4 & 0.5 & 0.6 & 0.7 & 0.8 & 0.9 & Avg. \\
    \midrule
    0 (global only) & 76 & 76 & 80 & 88 & 80 & 80 & 80.0 \\
    0.2 & 88 & 92 & 76 & 88 & 80 & 72 & 82.7 \\
    0.4 & 80 & 92 & \textbf{92} & 88 & \textbf{96} & 80 & 88.0 \\
    0.6 & 92 & 84 & 88 & 84 & 80 & \textbf{96} & 87.3 \\
    \oursrow
    0.8 (ours) & \textbf{100} & 96 & \textbf{92} & 92 & 88 & 92 & \textbf{93.3} \\
    0.9 & 92 & \textbf{100} & 88 & \textbf{96} & 92 & 72 & 90.0 \\
    1 (\moe{} only) & 56 & 48 & 48 & 24 & 20 & 32 & 38.0 \\
    \bottomrule
  \end{tabular}
\end{table}

\begin{figure}[!htb]
  \centering
  \includegraphics[width=0.6\textwidth]{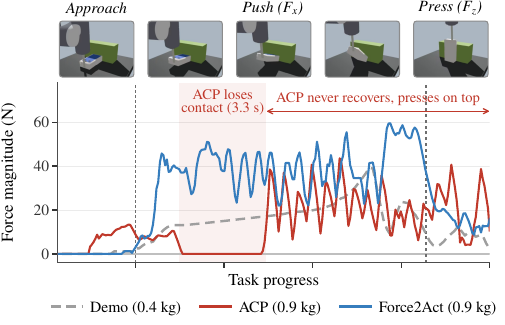}
  \caption{\textbf{Force during block pivoting at 0.9\,kg} against a 0.4\,kg
    demonstration, with both policies on the same seed.
    ACP loses contact and then only presses on top of the block, whereas Force2Act keeps the
    contact.
    The vertical dotted lines mark the first contact and where $F_z$ overtakes $F_x$.}
  \label{fig:force_profile}
\end{figure}

\cref{tab:pivot,tab:box,tab:round} report the per-condition results of block pivoting, press-push
box, and round-plug insertion that are averaged in \cref{tab:main}.
\cref{fig:force_profile} contrasts ACP and Force2Act on one rollout of block pivoting at 0.9\,kg.
\cref{tab:absforce} reports block pivoting per mass with the global force policy trained on
absolute force and with the MLP law in place of \moe{}.
On press-push box at $0^\circ$, $2^\circ$, and $4^\circ$, the MLP law succeeds in 9, 9, and 2 of
10 trials against 10 of 10 for Force2Act at every incline.
It presses with 6.3, 6.8, and 6.5\,N and Force2Act with 7.9, 10.8, and 15.0\,N, and they lose
contact in 2, 1, and 8 and in 1, 2, and 0 of 10 trials.

\begin{table}[!htb]
  \centering
  \footnotesize
  \caption{\textbf{Ablations on block pivoting per mass.}
    Successful trials, median contact force over all 25 trials, and trials that lose contact,
    out of 25 per mass (kg).
    A trial that never makes contact counts as 0\,N.
    The contact force is the mean of the measured $|F|$, smoothed over 5 steps, from the first
    contact ($|F| \ge 2$\,N on 3 consecutive steps) to the end of the trial, and a contact is
    lost when $|F| < 1$\,N after the first contact.
    \emph{Abs.} trains the global force policy on absolute force, \emph{MLP} replaces
    \moe{} with the MLP law at $w = 0.8$, and the last five rows are the design ablations of
    \cref{tab:design}.}
  \label{tab:absforce}
  \renewcommand{\arraystretch}{1.1}
  \setlength{\tabcolsep}{2.6pt}
  \begin{tabular}{l cccccc @{\hspace{10pt}} cccccc @{\hspace{10pt}} cccccc}
    \toprule
    & \multicolumn{6}{c}{Successful trials} & \multicolumn{6}{c}{Contact force (N)} & \multicolumn{6}{c}{Trials that lose contact} \\
    \cmidrule(lr){2-7} \cmidrule(lr){8-13} \cmidrule(lr){14-19}
    Mass (kg) & 0.4 & 0.5 & 0.6 & 0.7 & 0.8 & 0.9 & 0.4 & 0.5 & 0.6 & 0.7 & 0.8 & 0.9 & 0.4 & 0.5 & 0.6 & 0.7 & 0.8 & 0.9 \\
    \midrule
    \oursrow
    Force2Act & 25 & 24 & 23 & 23 & 22 & 23 & 25.6 & 28.9 & 30.1 & 31.4 & 33.2 & 34.3 & 5 & 6 & 7 & 6 & 5 & 4 \\
    Abs., $w = 0.8$ & 25 & 23 & 23 & 17 & 18 & 12 & 22.3 & 26.2 & 26.0 & 27.3 & 29.6 & 28.9 & 1 & 2 & 3 & 8 & 8 & 12 \\
    Abs., $w = 0$ & 10 & 14 & 6 & 10 & 4 & 0 & 7.0 & 13.6 & 6.1 & 10.9 & 6.5 & 6.6 & 17 & 17 & 20 & 17 & 25 & 24 \\
    MLP & 23 & 22 & 24 & 23 & 20 & 10 & 17.5 & 18.7 & 19.1 & 20.7 & 22.1 & 17.2 & 9 & 8 & 14 & 16 & 17 & 22 \\
    No bound & 22 & 25 & 25 & 23 & 24 & 16 & 24.9 & 24.2 & 25.1 & 25.7 & 26.3 & 26.7 & 8 & 3 & 4 & 5 & 4 & 10 \\
    No gates & 24 & 22 & 1 & 0 & 0 & 0 & 10.3 & 10.5 & 11.1 & 11.2 & 11.6 & 11.8 & 5 & 9 & 23 & 25 & 22 & 24 \\
    One per case & 23 & 22 & 24 & 22 & 18 & 16 & 24.5 & 24.9 & 26.2 & 28.5 & 29.4 & 31.3 & 5 & 7 & 13 & 7 & 12 & 10 \\
    Router sees dir. & 23 & 25 & 23 & 11 & 6 & 0 & 22.0 & 21.5 & 22.6 & 14.1 & 13.5 & 13.3 & 6 & 3 & 6 & 16 & 19 & 24 \\
    $\lambda = 0$ & 22 & 24 & 25 & 22 & 7 & 1 & 22.5 & 23.2 & 24.3 & 25.2 & 15.7 & 14.4 & 5 & 3 & 3 & 9 & 19 & 22 \\
    \bottomrule
  \end{tabular}
\end{table}

The overall success rate in \cref{tab:main,tab:ablation} pools all 305 trials
(\cref{tab:protocol}), so that each task is weighted by its number of trials.
\cref{tab:pivot_w} reports block pivoting per mass for every weight $w$ of \cref{tab:ablation}.

\section{Learned Laws and Router Analysis}
\label{sec:laws}

\begin{table}[!htb]
  \centering
  \footnotesize
  \caption{\textbf{Learned laws.}
    Constants of the \moe{} experts in~\eqref{eq:experts}, fitted jointly on the
    demonstrations of all five tasks, with forces in newtons.}
  \label{tab:laws}
  \renewcommand{\arraystretch}{1.1}
  \setlength{\tabcolsep}{4pt}
  \begin{tabular}{l l}
    \toprule
    Expert & Learned law \\
    \midrule
    Pushing & $F_n = 6.11$ \\
    Sticking & $F_n = 5.80\, F_t + 1.19$, \; $F_n = 7.51\, F_t + 1.97$ \\
    Sliding & $F_n \in \{5.99,\, 9.35,\, 16.14\}$ \\
    Insertion & $F_n = 1.55\, F_\perp + 7.73$ \\
    \bottomrule
  \end{tabular}
\end{table}

\paragraph{Learned laws.}
\cref{tab:laws} lists the laws that the single model learns from the five tasks.
The two sticking experts learn different slopes, 5.80 and 7.51, which are close to the
median ratios of the target normal force to the tangential force in the demonstrations of
press-push box (6.4) and block pivoting (8.4), and slightly lower, as the offsets carry part
of the force.
Bottle wiping, a sliding contact, has a ratio of only 4.0.
The gate of the second sticking expert admits most wiping frames, but this expert would
overestimate their normal force by about 6.7\,N.
Its squared error, about 46\,N$^2$, exceeds that of a sliding level by more than the penalty
$\lambda = 20$\,N$^2$, so the router assigns wiping to a sliding expert and thereby recovers
its contact mode without supervision.
The insertion expert adds 1.55\,N of axial push per newton of lateral force to a constant
7.73\,N, so the plug is pushed harder when it is loaded against the wall.

\paragraph{Experts and tasks.}
\cref{fig:real} shows the expert selected over each real rollout.
The experts are not the primitives of individual tasks.
No frame is labeled with its task, and each frame is assigned during training to the expert
that explains its normal force, so an expert describes a regime of force at the contact, and
a task may pass through several of them.
Block pivoting on the real robot illustrates this.
The tool presses the block against the wall at the 26\,N level of the third sliding expert,
drops to the 17\,N level of the second as the block tilts up, and returns to the third for the
last stage, in which it presses down with about 21\,N while moving about 9\,cm sideways.
This stage involves no wiping, yet a constant press combined with a lateral motion is the
form of a sliding expert.

\paragraph{Overlap between experts.}
Experts of different forms can also prescribe a similar force.
With little lateral force, the insertion law reduces to a nearly constant press, close to a
sliding level.
In one episode per task of Force2Act-Bench, round-plug insertion uses a sliding level of
about 6\,N on 85\% of its contact steps and the insertion law at about 8\,N on 9\%, and USB
insertion uses a level of about 9\,N on 24\% and the insertion law at about 10\,N on 76\%.
The router is least certain on round-plug insertion, with a top probability below 0.9 on
49\% of its contact steps, compared with at most 4\% on the other tasks of the benchmark,
which is consistent with this overlap.

\paragraph{Coverage.}
Where the force follows a load, the sticking experts dominate.
The first covers 88\% of press-push box in the benchmark, and 76\% of press-push box and 84\% of
press-push T on the real robot, and the second covers 54\% of block pivoting in the
benchmark, whereas a sliding expert covers 96\% of bottle wiping.
Outside these cases the router is close to one-hot, and its brief switches occur when the
robot pauses between action chunks, the tangential force falls to zero, and the gate of the
sticking expert closes.

\section{Design Ablations of \moe{} in Simulation}
\label{sec:design}

\cref{tab:design} in \cref{sec:exp:ablation} changes one part of \moe{} at a time, and
\cref{tab:absforce} gives block pivoting per mass.

\paragraph{Geometric gates.}
Without the gates, every expert is admissible on every frame, and the success rate falls to
48.5\%.
Bottle wiping succeeds in 1 of 50 trials, and on block pivoting the contact force stays
between 10.3 and 11.8\,N at every mass, so no trial succeeds from 0.7\,kg on.

\paragraph{One expert per contact case.}
Sharing one law between the sticking experts, $F_n = 6.64\, F_t + 1.89$, and one level of
8.69\,N between the sliding experts lowers the success rate to 78.0\%, with 22\% on bottle
wiping and 16 of 25 trials at 0.9\,kg.

\paragraph{Router input.}
When the router also observes the force direction, the contact force on
block pivoting drops from 22.6\,N at 0.6\,kg to 13.3 to 14.1\,N from 0.7\,kg on, and no trial
succeeds at 0.9\,kg, as anticipated in \cref{sec:method:routing}.
The success rate falls to 78.4\%.

\paragraph{Level penalty.}
With $\lambda = 0$, the laws fit the held-out demonstrations of block pivoting best among the
\moe{} variants, with an error of 6.75 against 7.90\,N, yet the success rate falls to 80.0\%.
The contact force drops to 15.7 and 14.4\,N at 0.8 and 0.9\,kg, where 7 and 1 of 25 trials
succeed, and to 1.2\,N on the $4^\circ$ incline, where 4 of 10 succeed.

\paragraph{Offset bound.}
Raising $b_{\max}$ from 2 to 60\,N, with all else unchanged, mainly changes the second
sticking expert.
Its law becomes $F_n = 6.14\, F_t + 7.54$ instead of $7.51\, F_t + 1.97$, while the first becomes
$6.80\, F_t + 0.65$ and the other experts change by at most 1.5\,N.
The offset then acts as a preload, and the contact force on block pivoting rises by only
1.8\,N from 0.4 to 0.9\,kg, against 8.7\,N for Force2Act (\cref{tab:absforce}).
Up to 0.8\,kg, this law succeeds in 119 of 125 trials against 117, but at 0.9\,kg it succeeds
in 16 of 25 against 23, so 135 of 150 trials succeed against 140.

\end{document}